\documentclass{article} %
\usepackage{iclr2027_conference,times}

\usepackage{amsmath,amsfonts,bm}

\def\eqref#1{equation~\ref{#1}}

\def\1{\bm{1}}

\DeclareMathAlphabet{\mathsfit}{\encodingdefault}{\sfdefault}{m}{sl}
\SetMathAlphabet{\mathsfit}{bold}{\encodingdefault}{\sfdefault}{bx}{n}

\usepackage{hyperref}
\usepackage{url}

\usepackage{lipsum}
\usepackage{booktabs}
\usepackage{multirow}
\usepackage{graphicx}
\usepackage[table]{xcolor}
\usepackage{bbm}
\usepackage{comment}

\usepackage{pifont}
\definecolor{checkgreen}{HTML}{4CAF50}

\newcommand{\cmark}{\textcolor{checkgreen}{\ding{51}}}
\newcommand{\xmark}{\textcolor{red}{\ding{55}}}
\usepackage{wrapfig}

\usepackage{xspace}

\newcommand{\oursmemupdater}{L3G-Mem\xspace}
\newcommand{\oursmem}{latent 3D Gaussian memory\xspace}
\newcommand{\oursvid}{L3G-Gen\xspace}
\newcommand{\ours}{L3G-World\xspace}
\newcommand{\paragrapht}[1]{\noindent\textbf{#1.}}

\usepackage{xstring,pgffor}

\definecolor{ppgA}{HTML}{60B5D7}
\definecolor{ppgB}{HTML}{70C482}

\newcommand{\gradtext}[1]{%
    \StrLen{#1}[\strlen]%
    \foreach \i in {1,...,\strlen} {%
        \StrChar{#1}{\i}[\thechar]%
        \pgfmathtruncatemacro{\pct}{100 - (\i-1)*100/(\strlen-1)}%
        \textcolor{ppgA!\pct!ppgB}{\thechar}%
    }%
}

\title{\ours: Real-Time Infinite World\\ Generation with Latent 3D Gaussian Memory\vspace{3mm}}

\author{
\begin{tabular}{l}
Jinhyeok Choi$^{1}$\thanks{Equal contribution.}
\quad
JoungBin Lee$^{1}$\footnotemark[1]
\quad
Jongmin Lee$^{1}$
\quad
Minkyeong Jeon$^{1}$
\quad
Jiyoung Kim$^{1}$
\\[1.0mm]
Kaeun Han$^{2}$
\quad
Yoonjin Yoon$^{1,3}$\thanks{Co-corresponding authors.}
\quad
Seungryong Kim$^{1}$\footnotemark[2]
\\[1.0mm]
{\normalfont\mdseries
$^{1}$KAIST AI
\qquad
$^{2}$Ewha Womans University
\qquad
$^{3}$KAIST Urban AI Institute}
\\[2mm]
{\normalfont\fontfamily{cmtt}\selectfont
\href{https://cvlab-kaist.github.io/L3G-World/}{
    \gradtext{https://cvlab-kaist.github.io/L3G-World/}
}}
\end{tabular}
}

\iclrfinalcopy %
\begin{document}

\maketitle
\lhead{L3G-World: Real-Time Infinite World Generation with Latent 3D Gaussian Memory}

\begin{figure}[h]
    \centering
    \vspace{-25pt}
    \includegraphics[width=0.9\linewidth]{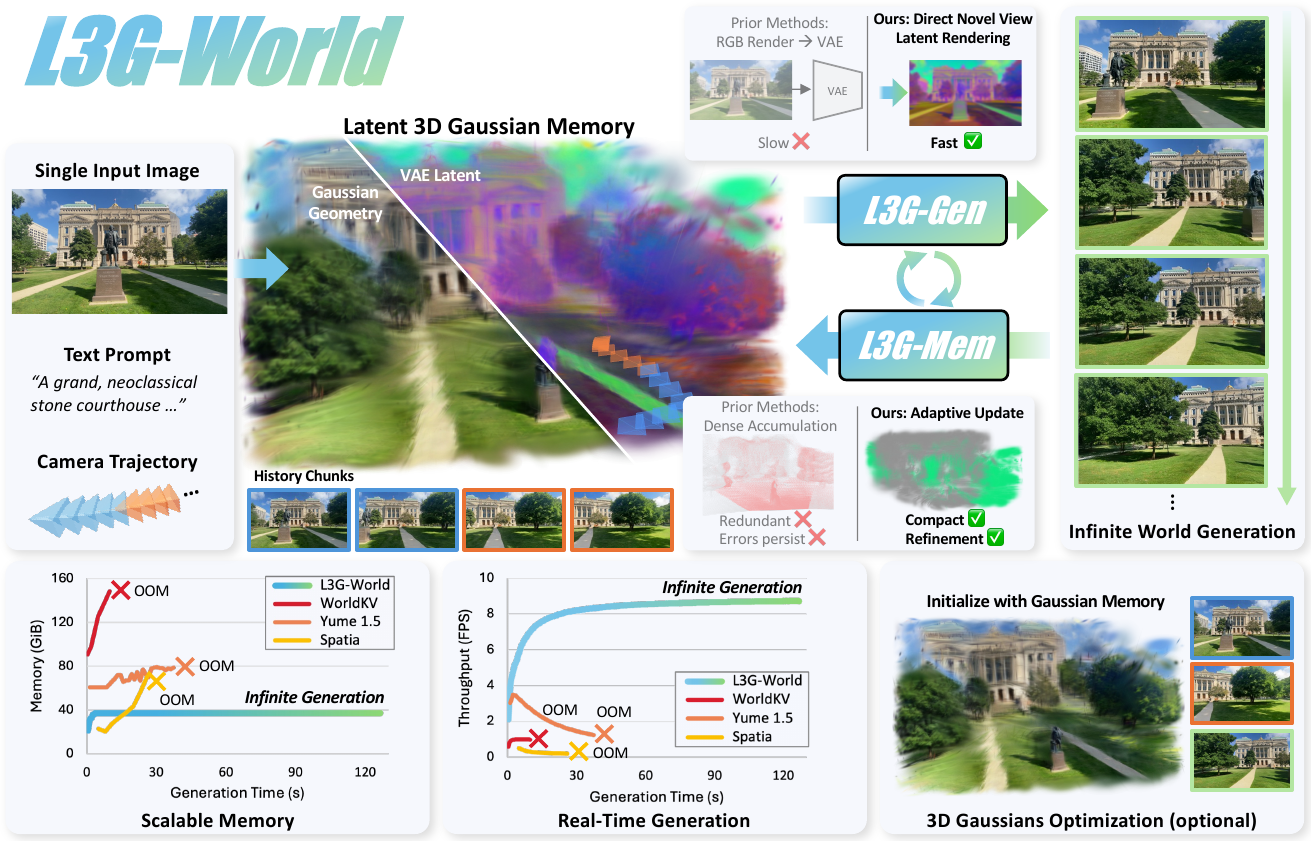}
    \vspace{-5pt}
    \caption{
    \textbf{Teaser.}
    \textbf{\ours} generates an explorable, spatially consistent world in real time from a single image and a camera trajectory.
    To this end, \oursmemupdater maintains a latent 3D Gaussian memory, and \oursvid generates video conditioned on VAE latents rendered from the memory.
    }
    \label{fig:teaser}
\end{figure}

\begin{abstract}

Interactive world models generate visual content conditioned on user actions, yet maintaining spatial consistency over long-horizon exploration in real time remains challenging.
Existing approaches rely on either temporal context or explicit 3D memory, both of which scale poorly because memory grows with the length of the generated stream. Moreover, explicit 3D memories condition latent video generators by re-encoding RGB renderings, which precludes real-time generation. In this paper, we introduce a \textbf{L}atent \textbf{3}D \textbf{G}aussian Memory (\textbf{L3G}) that combines the efficiency of feed-forward reconstruction with a compact, rapidly renderable scene representation. Building on this memory, we propose \textbf{\ours}, in which a memory updater and a video diffusion transformer (DiT) alternate autoregressively to generate spatially consistent worlds. The memory updater spawns Gaussians in newly observed regions and refines existing ones, so that the memory grows with scene extent rather than stream length. Each Gaussian further stores a VAE latent feature, allowing the memory to be rendered directly into the DiT's latent space without VAE re-encoding. Experiments show that \ours achieves state-of-the-art results on generation quality and long-horizon self-consistency, with up to $38.6\times$ fewer 3D primitives, $3.2\times$ faster memory reconstruction, and $359\times$ faster VAE feature rendering relative to the most efficient prior 3D-memory baseline for each metric, enabling real-time world generation at 9 FPS.

\end{abstract}
\section{Introduction}
Interactive video world models~\citep{seo2026grounding, nam2026worldcam, bruce2024genie, robbyantteam2026advancingopensourceworldmodels, mao2026yume1, sun2025worldplay, wang2026matrix, gao2026infinite, basant2026nvidia}, which generate visual content conditioned on user-provided actions, have recently emerged as a promising framework for real-world simulation~\citep{seo2026grounding, mao2026yume1, basant2026nvidia}, interactive game generation~\citep{bruce2024genie, robbyantteam2026advancingopensourceworldmodels, sun2025worldplay, wang2026matrix, gao2026infinite}, and embodied simulation~\citep{agarwal2026cosmos, chen2026abot}. A central challenge lies in maintaining spatial consistency during long-horizon exploration and loop closures while still supporting real-time generation.

One line of work employ temporal context, by retrieving history frames or KV caches of self-attention layers in diffusion models during autoregressive generation~\citep{robbyantteam2026advancingopensourceworldmodels, wang2026matrix,mao2026yume1,yi2026worldkvefficientworldmemory}. By allowing each new chunk to attend to information from earlier generations through attention, these methods help maintain scene consistency~\citep{cameo_Kwon_2026_CVPR}. However, its memory and attention costs grows with the length of the generated stream and spatial resolution. As shown in Tab.~\ref{table:contribution_of_memory}, this linear growth makes long-horizon generation increasingly expensive and can eventually lead to out-of-memory failures.

Another line of work explicitly stores scene information in persistent 3D representations. These approaches preserve previously generated content by reconstructing frames into point-cloud-based memories~\citep{ren2025gen3c, zhao2026spatia, wu2026spmem, shen2026lyra, lee20263d} or 3D Gaussian primitives~\citep{de2026artifixer, wu2025difix3d+, yang2026neoverse}. Such spatially anchored representations improve consistency across revisited regions while enabling fine-grained camera control. However, as shown in Tab.~\ref{table:contribution_of_memory}, existing approaches exhibit a fundamental trade-off between reconstruction efficiency and throughput. Feed-forward methods based on per-pixel point clouds or Gaussian primitives enable fast scene reconstruction, but predicting dense 3D representations for every pixel introduces substantial memory overhead. As generation proceeds, these dense primitives are repeatedly accumulated over the same spatial regions, resulting in increasingly redundant scene representations and slower target-view rendering. Moreover, since previously stored primitives are not adaptively updated, geometrically inaccurate regions may persist over time. In contrast, optimization-based methods~\citep{de2026artifixer, wu2025difix3d+} maintain more compact representations whose memory footprint primarily scales with scene extent, but their iterative optimization introduces substantial latency, making them unsuitable for real-time generation.

To address these challenges, we introduce a compact {\textbf{L}atent \textbf{3}D \textbf{G}aussian Memory }(\textbf{L3G}), which is continually read and updated online, and whose size grows primarily with the spatial extent of the explored scene rather than with the length of the generated stream. Building on this memory, we propose \textbf{\ours}, a world-generation framework composed of two modules: a feed-forward memory updater \textbf{\oursmemupdater}, and a memory-conditioned video diffusion transformer \textbf{\oursvid}.
As the two modules alternate autoregressively, \oursmemupdater spawns new Gaussians only in uncovered regions, and refines the existing Gaussians in explored regions through residual updates. The memory is therefore not a buffer that accumulates primitives, but a persistent scene representation revised over time, which avoids redundant accumulation and progressively corrects geometrically inaccurate Gaussians. Since each Gaussian additionally stores a VAE latent feature, the memory is rendered directly into the latent space of \oursvid under the target cameras, conditioning the generation of the next video chunk without VAE re-encoding. The generated chunk is then fed back to \oursmemupdater, which again updates the memory, closing the autoregressive loop. Combined with few-step distillation of \oursvid, this design enables real-time, long-horizon world generation.

Experiments demonstrate that \ours achieves state-of-the-art long-horizon consistency and generation quality while using $26.7$--$38.9\times$ fewer 3D primitives, performing memory reconstruction $3.0$--$3.2\times$ faster, and providing $349$--$359\times$ higher VAE-space rendering throughput than the most efficient prior explicit 3D-memory baseline for each metric. Combined with few-step distillation of \oursvid, this enables real-time world generation at 9.02 FPS on one H100 GPU.

\section{Related Works}
\begin{table*}[t]
\centering
\caption{\textbf{Computational efficiency of memory analysis.}
We compare memory scaling across methods and report memory and throughput for constructing and rendering explicit scene representations.
For memory scaling, $T$ denotes the number of generated frames, $R$ denotes the resolution size, and $S$ denotes the spatial coverage of the scene. `OOM' denotes out-of-memory error.
All cost analyses are measured by 400-frame video generation on one H100 GPU.
}
\label{table:contribution_of_memory}
\resizebox{\textwidth}{!}{
\begin{tabular}{l c c c c c c c c c c}
\toprule
Method
& Resolution
& Memory Type
& Memory Build
& Refinement
& \shortstack{Memory\\Scaling $\downarrow$}
& \shortstack{Peak GPU\\Memory$\downarrow$}
& \shortstack{Total\\FPS$\uparrow$}
& Real-time \\
\midrule

WorldKV~\citep{yi2026worldkvefficientworldmemory}
& $464 \times 832$
& Temporal context
& KV cache
& --
& $\mathcal{O}(TR)$
& OOM
& OOM
& \xmark \\

Yume 1.5~\citep{mao2026yume1}
& $704 \times 1,280$
& Temporal context
& Frame compression
& --
& $\mathcal{O}(TR)$
& 66.7 GiB
& 1.42
& \xmark \\

\midrule

Spatia~\citep{zhao2026spatia}
& $704 \times 1,248$
& Point clouds
& Feed-forward
& \xmark
& $\mathcal{O}(TR)$
& 32.6 GiB
& 0.24
& \xmark \\

NeoVerse~\citep{yang2026neoverse}
& $336 \times 560$
& 3D Gaussians
& Feed-forward
& \xmark
& $\mathcal{O}(TR)$
& 36.8 GiB
& 1.99
& \xmark \\

ArtiFixer~\citep{de2026artifixer}
& $540 \times 960$
& 3D Gaussians
& Optimization
& \cmark
& $\mathcal{O}(S)$
& 11.7 GiB
& 0.10
& \xmark \\

\midrule

\textbf{\ours (Ours)} 
& $480 \times 832$
& \textbf{Latent 3D Gaussians}
& \textbf{Feed-forward}
& \cmark
& $\mathcal{O}(S)$
& \textbf{7.4 GiB}
& \textbf{9.02}
& \cmark \\

\bottomrule
\end{tabular}
}
\vspace{-10pt}
\end{table*}

\paragrapht{Video Diffusion and World Models}
Video diffusion models have made substantial progress in high-quality video generation~\citep{yang2025cogvideox, wan2025wan, ali2025world, agarwal2026cosmos, hacohen2026ltx}.
Most adopt latent diffusion~\citep{rombach2021highresolution, blattmann2023videoldm}, where raw videos are compressed into a latent space using a VAE.
Reducing the number of sampling steps, together with causal generation, has further enabled low-latency video streaming~\citep{yin2025causvid, huang2026self, lin2026autoregressive, nie2026transition, gu2026anyflow}.
They improve few-step generation by training on model-generated histories to mitigate exposure bias, or by combining flow-based transitions with on-policy distillation.
Building on these advances, interactive world models~\citep{seo2026grounding, nam2026worldcam, bruce2024genie, robbyantteam2026advancingopensourceworldmodels, mao2026yume1, sun2025worldplay, wang2026matrix, gao2026infinite, basant2026nvidia, yi2025deep} generate interactive environments conditioned on user or agent actions.

\paragrapht{Feed-forward 3D Reconstruction}
Traditional reconstruction methods~\citep{mildenhall2020nerf, kerbl3Dgaussians} rely on per-scene optimization, whereas recent feed-forward approaches directly predict 3D geometry from input images in a single forward pass.
They typically predict per-pixel pointmap~\citep{wang2025vggt, da3} or 3D Gaussians~\citep{charatan2024pixelsplat, chen2024mvsplat, xu2025depthsplat}.
Recent works~\citep{wang2025cut3r, zhuo2025streamvggt, li2025wint3rwindowbasedstreamingreconstruction, chen2025ttt3r} further introduce streaming formulations that sequentially process incoming images for online reconstruction.
Complementarily, token-based 3DGS methods~\citep{an2025c3g, veicht2026zipsplat, tokengs2026, yoo2026instok3d} decode compact sets of Gaussian primitives from visual or learnable tokens rather than predicting dense per-pixel Gaussians.

\paragrapht{Consistent World Generation}
To improve long-term spatial consistency, several world models maintain explicit 3D scene memories, typically represented as point clouds~\citep{ren2025gen3c, zhao2026spatia, wu2026spmem, shen2026lyra, lee20263d} or 3D Gaussians~\citep{yang2026neoverse, de2026artifixer, wu2025difix3d+}.
These methods reconstruct 3D representations from generated frames and render them into subsequent target views to provide spatially grounded conditioning.
Such explicit 3D memories improve consistency and camera controllability, but feed-forward approaches typically rely on dense per-frame reconstruction and continuously accumulate new 3D primitives, causing memory consumption and rendering cost to grow with the generation horizon.
Some methods~\citep{de2026artifixer, wu2025difix3d+} use optimization-based reconstruction to maintain a scene-level 3D representation rather than simply accumulating per-frame primitives, but their slow per-scene optimization introduces substantial latency and is difficult to reconcile with real-time generation.
In contrast, \ours maintains compact 3D Gaussians as scene memory, by adaptively spawning new primitives or refining existing ones.
Moreover, each Gaussian directly stores VAE latent features, enabling efficient conditioning of the video diffusion model for real-time world generation.

\section{Methodology}
\label{sec:method}

\begin{figure}[t]
    \centering
    \vspace{-10pt}
    \includegraphics[width=1\linewidth]{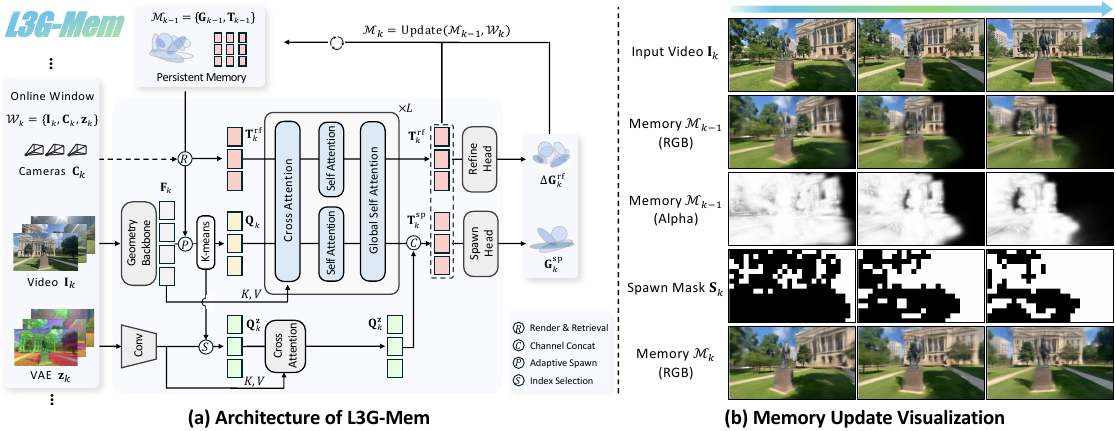}
    \vspace{-15pt}
    \caption{
       \textbf{\oursmemupdater.} 
    (a) \oursmemupdater takes the current window $\mathcal W_k$ and the memory $\mathcal M_{k-1}$. Rendering $\mathcal M_{k-1}$ into the current cameras determines which geometry features are clustered by K-means into spawning queries, and identifies the visible Gaussians whose memory tokens are retrieved as refinement tokens. Both token sets are jointly processed by a transformer and decoded into new Gaussians and residual updates, respectively.
    (b) Given the input video chunk, RGB and opacity rendered from the previous memory identify regions requiring new Gaussians through a spawning mask. As a result, the updated memory can fully reconstruct the observed scene.
    }
    \vspace{-5pt}
    \label{fig:gs_model_archi}
\end{figure}

\subsection{Motivation and Overview}
Real-time interactive world generation that maintains geometric consistency places two competing demands on a 3D memory.
Consistency over a long horizon requires storing previously generated content, while keeping pace with generation requires a memory that stays compact and fast to read and write. Dense per-frame accumulation stores everything but grows unboundedly, since redundant primitives pile up over the same regions and slow down rendering.
We adopt 3D Gaussians as the memory representation for their robust target-view rendering, and design \oursmem around two decisions. First, the memory is updated selectively: it \emph{spawns} new Gaussians only for newly observed geometry and \emph{refines} existing ones to correct reconstruction errors from earlier steps, so that memory usage remains largely independent of the number of generated frames. Second, each Gaussian stores a VAE latent feature, so that the memory is rendered directly in the latent space of the video generator and conditions the generator without VAE re-encoding. 

As illustrated in Fig.~\ref{fig:gs_model_archi}, \ours realizes these decisions with two modules that alternate autoregressively: a memory updater, \oursmemupdater, which reads and updates the \oursmem with each generated chunk, and a video generator, \oursvid, which generates the next chunk conditioned on the memory rendered in its latent space.
We first present \oursmemupdater and the \oursmem (Sec.~\ref{sec:method_gaussian}), then describe how \oursvid is conditioned on the memory and distilled for real-time inference (Sec.~\ref{sec:method_diffusion}), and finally provide the world generation framework \ours (Sec.~\ref{sec:method_framework}).
Implementation details are provided in Sec.~\ref{sec:appendix_impl_gs} and Sec.~\ref{sec:appendix_impl_diffusion}, respectively.

\subsection{\oursmemupdater: 3D Gaussians as World Memory}
\label{sec:method_gaussian}

\paragrapht{Online Window-Based Memory Update}
\oursmemupdater processes a stream of images $\mathrm{I}^i \in \mathbb{R}^{3\times H \times W}$ and their corresponding camera parameters $\mathrm{C}^i$ in windows of size $V$.
At autoregressive step $k$, we define the input window as
$\mathcal{W}_k = \{(\mathrm{I}^i_k, \mathrm{C}^i_k, z^i_k)\}_{i=1}^{V} = \{\mathbf{I}_k, \mathbf{C}_k, \mathbf{z}_k \}$,
where $\mathbf z_k$ denotes the VAE latent of $\mathbf I_k$.
The \oursmem after processing this window is represented as:
\begin{equation}
    \mathcal{M}_k = \left\{(\mathrm{G}_j, \mathrm{T}_j)\right\}_{j=1}^{N_k} = \{\mathbf{G}_k , \mathbf{T}_k \},
\end{equation}
where $\mathrm{T}_j$ denotes a memory token,
$\mathrm{G}_j$ is the corresponding 3D Gaussians,
and $N_k$ is the number of memory tokens after processing the $k$-th window.
The Gaussian parameters are decoded from each memory token using an MLP-based Gaussian head,
$\mathrm{G}_j = \operatorname{Head}(\mathrm{T}_j)$.

Given the previous memory $\mathcal{M}_{k-1}$ and the current input window $\mathcal{W}_k$,
our model updates the memory autoregressively as:
\begin{equation}
    \mathcal{M}_k
    =
    \operatorname{Update}
    \left(\mathcal{M}_{k-1}, \mathcal{W}_k\right).
\end{equation}

\paragrapht{Adaptive Gaussian Spawning}
To keep \oursmem compact, \oursmemupdater selectively spawns new Gaussians only where additional scene geometry is needed. This requires identifying regions that are insufficiently represented by the current memory and initializing spawning query tokens from those regions, leaving well-represented regions to the existing Gaussians. To identify such regions, we render the Gaussians stored in $\mathcal M_{k-1}$ from the current camera views $\mathbf C_k$ to obtain RGB images and opacity maps. The opacity indicates whether each pixel is covered by the memory, and the per-pixel RGB error against $\mathbf I_k$ indicates whether the memory reproduces that pixel faithfully. We mark pixels with low opacity or high error as insufficiently represented, and collect them into a spawn mask $\mathbf S_k$.

The spawn mask then determines where new Gaussians are spawned: we draw the spawning queries from the masked observation, so that the new Gaussians fill the regions marked by $\mathbf S_k$. Specifically, we encode $\mathbf I_k$ with a geometric foundation model~\citep{da3} to obtain geometry features $\mathbf F_k{=}\{\mathrm F_k^l\}_{l=1}^{L}$, where $L{=}3$ denotes the selected feature layers, and retain only the features of $\mathrm F_k^1$ inside $\mathbf S_k$ as spawning candidates. Following ZipSplat~\citep{veicht2026zipsplat}, we apply K-means to these candidates to initialize a compact set of spawning query tokens $\mathbf Q_k$. Thus, rather than clustering over all input geometry features, our updater allocates its spawning budget only to content missing from the memory.

\paragrapht{Gaussian Refinement}
The refinement stream of \oursmemupdater updates existing Gaussians that are visible in the current window, rather than leaving previously stored content fixed.
We render the Gaussians stored in $\mathcal{M}_{k-1}$ from the current camera views $\mathbf{C}_k$, retrieve the memory tokens associated with the visible Gaussians, and use them as refinement tokens, denoted by $\mathbf{T}_k^{\mathrm{rf}}$.
These tokens and their corresponding Gaussian parameters are subsequently updated using observations from the current window $\mathcal{W}_k$.

\paragrapht{Joint Token Processing}
\oursmemupdater concatenates the spawning and refinement tokens along the token dimension and processes them with a single transformer $\mathcal{F}_\theta$:
\begin{equation}
    \left(\bar{\mathbf{Q}}_k, \bar{\mathbf{T}}_k^{\mathrm{rf}}\right)
    =
    \mathcal{F}_\theta\left(
        \left[\mathbf{Q}_k; \mathbf{T}_k^{\mathrm{rf}}\right],
        \mathbf F_k
    \right),
\end{equation}
where $[\cdot;\cdot]$ denotes concatenation along the token dimension. The transformer consists of $L$ blocks.
In each block $l$, the two streams first cross-attend to the geometric features $\mathrm F_k^l$, followed by stream-wise self-attention.
A global self-attention layer is then applied over the concatenated spawning query and refinement tokens, enabling information exchange between newly observed and previously reconstructed scene content.

\paragrapht{VAE Latent Lifting}
Since video diffusion models operate in the VAE latent space, we additionally lift VAE features into \oursmem, allowing it to be rendered directly in the latent space of the diffusion model.
A dedicated latent-lifting branch initializes latent queries $\mathbf Q^\mathbf z_k$ from $\mathbf z_k$ using the same spatial indices as the spawning query tokens, and cross-attends to $\mathbf z_k$.
This produces VAE latent tokens that are spatially aligned with the corresponding query tokens, and can therefore be jointly decoded as $\mathbf T^\mathrm{sp}_k=\operatorname{Concat}_c(\bar{\mathbf{Q}}_k;\bar{\mathbf Q}^\mathbf{z}_k)$, where $\operatorname{Concat}_c(\cdot;\cdot)$ is channel-wise concatenation.

\paragrapht{Gaussian Decoding and Memory Update}
The processed spawning tokens are concatenated with their corresponding VAE latent tokens and decoded into new Gaussian primitives, while the refinement tokens are decoded into residual updates for the retrieved Gaussians:
\begin{equation}
    \mathbf{G}_k^{\mathrm{sp}}
    =
    \operatorname{Head}_{\mathrm{sp}}
    \left(
        \mathbf T^\text{sp}_k
    \right),
    \qquad
    \Delta\mathbf{G}_k^{\mathrm{rf}}
    =
    \operatorname{Head}_{\mathrm{rf}}
    \left(
        \bar{\mathbf{T}}_k^{\mathrm{rf}}
    \right),
\end{equation}
Where $\operatorname{Head}_{\operatorname{sp}}(\cdot)$ and $\operatorname{Head}_{\operatorname{rf}}(\cdot)$ are MLP-based Gaussian head for spawn and refine tokens, respectively.
To obtain $\mathcal{M}_k$, \oursmemupdater updates $\mathcal{M}_{k-1}$ in two ways. It appends the newly spawned Gaussians together with their memory tokens, and updates the retrieved Gaussians in place by applying the predicted residuals and replacing their memory tokens with the refinement outputs.
All memory entries not selected for refinement remain unchanged.

\paragrapht{Autoregressive Training}
We train over $K$ consecutive windows, starting from an empty memory $\mathcal M_0= \emptyset$. We apply stop gradients through $\mathcal{M}_{k-1}$ at each step $k$ to ensure causality.
At step $k$, we supervise renderings of $\mathcal{M}_k$ from the target views of all windows up to $k$, with the overall objective
\begin{equation}
    \label{eq:reconloss}
    \mathcal{L}
    =
    \frac{1}{K}\sum_{k=1}^{K}
    \left(
        \mathcal{L}_{\mathrm{rgb}}^k
        + \lambda_{\mathrm{lpips}}\mathcal{L}_{\mathrm{lpips}}^k
        + \lambda_{\mathrm{depth}}\mathcal{L}_{\mathrm{depth}}^k
        + \lambda_{\mathrm{vae}}\mathcal{L}_{\mathrm{vae}}^k
        + \lambda_{\mathrm{vis}}\mathcal{L}_{\mathrm{vis}}^k
    \right).
\end{equation}
The terms denote RGB $\ell_1$, LPIPS, depth $\ell_1$, VAE latent MSE, and visibility losses, respectively.
The visibility loss~\citep{tokengs2026} encourages each Gaussian to lie within at least one input-view frustum.

\begin{figure}[h]
    \centering
   \vspace{-10pt}
   \includegraphics[width=\linewidth]{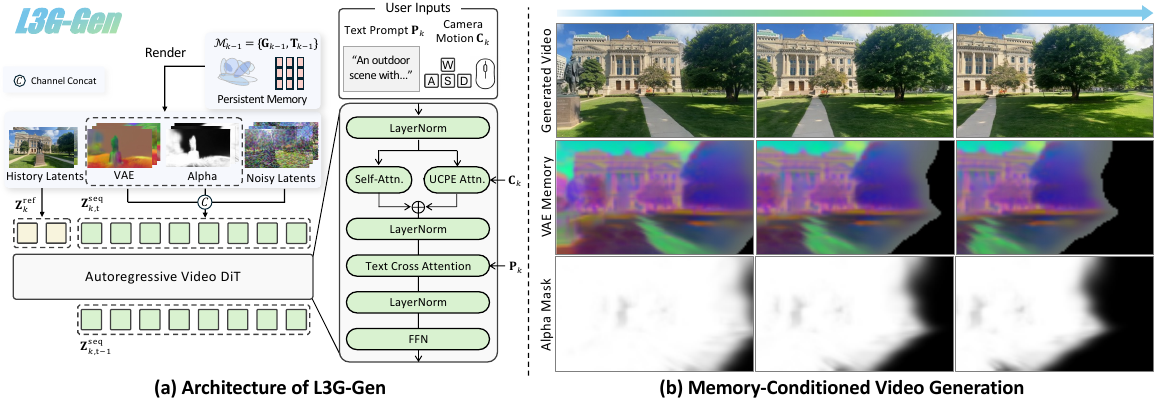}
    \vspace{-15pt}
    \caption{
    \textbf{\oursvid.} (a) \oursvid renders the VAE latents and opacity of the 3D memory to the target cameras, concatenates them with the noisy latents along the channel dimension, and generates the next chunk conditioned on the history latents.
    (b) We visualize the ground-truth and generated video, with target-view rendered VAE condition from the \oursmem, illustrating that spatially aligned latent memory provides consistent conditioning for video generation.
    }
    \label{fig:dit_framework}
\end{figure}

\subsection{\oursvid: Memory-conditioned Autoregressive Video Generation}
\label{sec:method_diffusion}
We introduce \oursvid, a memory-conditioned autoregressive video diffusion model for long-horizon world generation.
At each autoregressive step, the model retrieves previously generated information from the persistent 3D Gaussian memory to guide the generation of the next video chunk.
This provides spatially aligned conditioning for preserving scene consistency when revisiting previously explored regions.
To support real-time interaction, we further distill the model into a few-step generator.

\paragrapht{Conditioning for Generation}
At autoregressive step $k$, we render the VAE latents stored in $\mathcal{M}_{k-1}$ under the target cameras $\mathbf{C}_k$ to obtain spatially aligned VAE latents ${\mathbf{z}}_k^\mathrm{mem}$ and alpha mask $\mathbf o_k^\text{mem}$. We concatenate them with the target video latents $\mathbf{z}_{k,t}$ along the channel dimension, providing geometric conditioning without introducing additional memory tokens.
The previously generated reference latents $\mathbf{z}_k^{\mathrm{ref}}$ are concatenated with zero-filled channels to match the channel dimension of the concatenated target latents. The input sequence is formed by token-wise concatenation:
\begin{equation}
    \mathbf{z}_{k,t}^{\mathrm{seq}}
    =
    \left[
        \mathbf{z}_k^{\mathrm{ref}};
        \operatorname{Concat}_c(\mathbf{z}_{k,t};{\mathbf{z}}_k^\text{mem};\mathbf o_k^\text{mem})
    \right].
    \label{eq:latent_sequence}
\end{equation}
Only the target video latents are denoised, while the reference latents remain fixed. 
Following ~\citep{zhang2026unified}, we add the residual camera ray rotary positional embedding (RayRoPE~\citep{wu2026rayrope}) on the self attention in each DiT layer.

\paragrapht{Temporal Positional Encoding}
Following~\citep{seo2026grounding}, we assign the reference latents a temporally offset rotary positional embedding~\citep{su2021roformer} (RoPE) preceding the current generation chunk.
The temporal RoPE positions associated with the input sequence are
\begin{equation}
    \mathbf{p}_k^{\mathrm{seq}}
    =
    \left[
        \underbrace{-L_\text{ref},\ldots,-1}_{\text{reference}}; \
        \underbrace{0,\ldots,L{-}1}_{\text{target}}
    \right],
    \label{eq:temporal_rope}
\end{equation}
where $L_\text{ref}$ and $L$ denote the numbers of reference and target latents, respectively.
All spatial tokens within each latent frame share its temporal RoPE position.
The reference latents serve as conditioning rather than a future prediction target, and remain accessible to all target tokens.

\paragrapht{Training Objective}
We train the memory-conditioned teacher model with rectified flow~\citep{lipman2022flow}.
Let $\mathbf{z}_{k,0}$ denote the clean target video latents and
$\mathbf{z}_{k,t}=(1-t)\mathbf{z}_{k,0}+t\boldsymbol{\epsilon}$ is noisy target latents,
where $\boldsymbol{\epsilon}\sim\mathcal{N}(0,\mathbf{I})$ is a sampled noise and
$t\sim\mathcal{U}(0,1)$ is a noise level.
Conditioned on the rendered memory latents
${\mathbf{z}}_k^{\mathrm{mem}}$, opacity map $\mathbf o_k^\text{mem}$, retrieved reference latents
$\mathbf{z}_k^{\mathrm{ref}}$, target cameras $\mathbf{C}_k$, and the text prompt $\mathbf P_k$
the teacher $\mathbf{v}_{\phi}$ is optimized as
\begin{equation}
    \mathcal{L}_{\mathrm{FM}}
    =
    \mathbb{E}_{\mathbf{z}_{k,0},{\epsilon},t}
    \left[
        \left\|
        \mathbf{v}_{\phi}
        \!\left(
            \mathbf{z}_{k,t},t
            \mid
            {\mathbf{z}}_k^{\mathrm{mem}},
            \mathbf o_k^\text{mem},
            \mathbf{z}_k^{\mathrm{ref}},
            \mathbf{C}_k,
            \mathbf P_k
        \right)
        -
        ({\epsilon}-\mathbf{z}_{k,0})
        \right\|_2^2
    \right].
    \label{eq:conditional_flow_matching}
\end{equation}
The loss is applied only to the target video latents, while the clean reference latents remain fixed conditioning inputs.

\paragrapht{On-Policy Distillation}
To reduce the sampling cost, we distill the memory-conditioned teacher $\mathbf{v}_{\phi}$ into a few-step autoregressive student following AnyFlow~\citep{gu2026anyflow}.
The student is conditioned on both the source and destination noise levels and learns flow-map transitions between arbitrary points along the denoising trajectory.
After flow-map initialization, we jointly optimize the flow-map objective and on-policy distribution matching.
Specifically, the student generates samples through its own autoregressive rollout, which are re-noised and compared against the frozen teacher through distribution matching.
Training on the student's own generation trajectory, rather than relying solely on noisy ground-truth samples, reduces the train--test mismatch in few-step autoregressive generation.
Detailed distillation objectives and training configurations are provided in Sec.~\ref{sec:prelim_diffusion}.

\subsection{\ours: Consistent World Generation Framework}
\label{sec:method_framework}

\begin{wrapfigure}{r}{0.55\textwidth}
    \centering
    \vspace{-10pt}
    \includegraphics[width=1\linewidth]{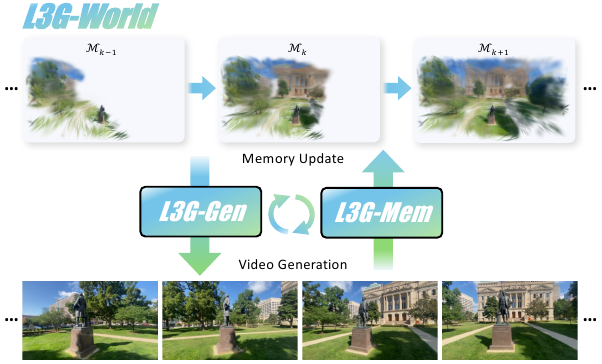}
    \vspace{-15pt}
    \caption{
    \textbf{\ours Pipeline.} \ours couples video generation and memory updating in a chunk-wise loop, where each generated chunk updates the persistent 3D memory, which in turn conditions subsequent chunk generation.
    }
    \vspace{-10pt}
    \label{fig:archi_pipeline}
\end{wrapfigure}

Combining \oursmemupdater and \oursvid, \ours performs world generation through an autoregressive loop of adaptive memory update and memory-conditioned video generation.
Given an input image, \oursmemupdater first initializes the persistent Gaussian memory $\mathcal M_0$.
At each generation step $k$, the VAE features stored in $\mathcal M_{k-1}$ are rendered from the target cameras $\mathbf C_k$ to provide spatially aligned latent conditioning for \oursvid.
Importantly, we directly render memory in the VAE latent space, avoiding RGB rendering followed by VAE re-encoding.

Conditioned on the rendered memory, \oursvid generates the next video chunk in the VAE latent space.
These generated latents are passed directly to \oursmemupdater for VAE feature lifting, while the corresponding RGB frames are obtained once through the VAE decoder.
The updater then update the memory, producing $\mathcal M_k$ for the next generation step.
By repeatedly reading from and writing back to the same persistent 3D memory, \ours continuously extends and corrects the generated world while avoiding redundant latent encoding between the two modules. \ours Fast employ few-step student model of \oursvid, enabling real-time world generation.
\section{Experiments}

\subsection{Experimental Setup}

\paragrapht{Implementation Details}
We implement \oursmemupdater inspired by ZipSplat~\citep{veicht2026zipsplat}, using Depth Anything 3 (DA3)~\citep{da3} as the geometry backbone. For \oursvid, we adopt Cosmos-Predict2.5~\citep{nvidia2025cosmos2.5}. We train \oursmemupdater on DL3DV~\citep{ling2024dl3dv}, and \oursvid on DL3DV and RealEstate10K~\citep{zhou2018stereomagnificationlearningview}. We fine-tune Diffusion Transformer (DiT) backbone of the video generator using LoRA~\citep{hu2022lora}, and we adopt AnyFlow~\citep{gu2026anyflow} for few-step distillation. We refer to the few-step distilled variant as `\ours Fast' in our experiments. All models are optimized using AdamW~\citep{loshchilov2017decoupled} on four NVIDIA A100 GPUs, with a batch size of 8 and 16 for the memory updater and world model, respectively. The implementation and training details of the memory updater and video DiT are provided in Sec.~\ref{sec:appendix_impl_gs} and ~\ref{sec:appendix_impl_diffusion}.

\paragrapht{Baselines}
We compare \ours against state-of-the-art world models with either explicit or implicit memory, including point-cloud-based methods, Lyra 2.0~\citep{shen2026lyra}, GEN3C~\citep{ren2025gen3c}, Spatia~\citep{zhao2026spatia}, SPMem~\citep{wu2026spmem}, and 3DScenePrompt~\citep{lee20263d}; the Gaussian-based method NeoVerse~\citep{yang2026neoverse}; and temporal-context-based methods, Yume 1.5~\citep{mao2026yume1} and Matrix-Game 3.0~\citep{wang2026matrix}. For a fair comparison in our single-image setting, we use Depth Anything 3 GIANT-1.1-NESTED~\citep{da3} to obtain metric depth for all explicit 3D-memory methods, aligning the scale of the reconstructed scene representation with the evaluation camera trajectory. For Gaussian-based methods, including NeoVerse and \ours, we align the initially predicted Gaussian geometry to the same metric depth. All methods are evaluated on a single NVIDIA A100 GPU, including the efficiency measurements. Detailed baseline configurations are provided in Sec.~\ref{sec:appendix_baseline}.

\paragrapht{Evaluation Benchmark}
We evaluate \ours and all baselines on DL3DV~\citep{ling2024dl3dv} and RealEstate10K~\citep{zhou2018stereomagnificationlearningview}. To evaluate long-horizon spatial consistency, we construct 481-frame loop-closure trajectories that explore away from the initial view and then retrace the same camera path in reverse. We measure \textit{self-consistency} by comparing generated frames at corresponding camera poses along the outgoing and returning trajectories using PSNR, SSIM, and LPIPS.
For generation quality, we report LPIPS, FID~\citep{heusel2017gans}, and four VBench~\citep{huang2023vbench} metrics: Imaging Quality (IQ), Aesthetic Quality (AQ), Subject Consistency (SC), and Background Consistency (BC). Following~\citep{seo2026grounding}, we evaluate camera controllability using translation error (TransErr) and rotation error (RotErr), and evaluate geometric consistency using MEt3R~\citep{asim25met3r}, following~\citep{jang2026repurposing}.

To evaluate the efficiency of 3D memory, we measure the number of stored primitives, memory footprint, reconstruction and rendering FPS. We separately report RGB rendering FPS and VAE-space conditioning FPS, where the latter includes VAE encoding after RGB rendering for methods without directly renderable VAE features. Further evaluation details are provided in Sec.~\ref{sec:appendix_eval}.

\begin{table*}[t]
\centering
\caption{
\textbf{Video generation evaluation.}
We evaluate generation quality, self-consistency, and efficiency on DL3DV and RealEstate10K.
The best and second-best results for each metric within each dataset are shown in
\textbf{bold} and \underline{underline}, respectively.
}
\label{tab:main_eval_video_l3g_world}
\resizebox{\textwidth}{!}{
\begin{tabular}{lccccc|ccc|cccccc|cc|c}
\toprule
& \multicolumn{5}{c}{\textbf{3D Memory Efficiency}}
& \multicolumn{3}{c}{\textbf{Self-Consistency}}
& \multicolumn{6}{c}{\textbf{Generation Quality}}
& \multicolumn{2}{c}{\textbf{Pose Error}}
& \multicolumn{1}{c}{\textbf{3D Consistency}} \\
\cmidrule(lr){2-6}
\cmidrule(lr){7-9}
\cmidrule(lr){10-15}
\cmidrule(lr){16-17}
\cmidrule(lr){18-18}

Method
& \shortstack{\# Primitives\\(K) $\downarrow$}
& \shortstack{Memory\\(MiB) $\downarrow$}
& \shortstack{Recon. FPS\\$\uparrow$}
& \shortstack{Render FPS\\(RGB) $\uparrow$}
& \shortstack{Render FPS\\(VAE) $\uparrow$}
& PSNR $\uparrow$
& SSIM $\uparrow$
& LPIPS $\downarrow$
& LPIPS $\downarrow$
& FID $\downarrow$
& IQ $\uparrow$
& AQ $\uparrow$
& SC $\uparrow$
& BC $\uparrow$
& TransErr $\downarrow$
& RotErr ($^\circ$) $\downarrow$
& MEt3R $\downarrow$ \\
\midrule

\multicolumn{18}{c}{\textit{DL3DV}} \\
\midrule

\multicolumn{18}{l}{\textit{Temporal Context}} \\

Yume 1.5
& --
& --
& --
& --
& --
& 13.71
& 0.487
& 0.655
& 0.699
& 50.88
& 0.601
& \underline{0.514}
& 0.805
& 0.888
& 0.904
& 11.09
& \textbf{0.050} \\

Matrix Game 3.0
& --
& --
& --
& --
& --
& 12.26
& 0.401
& 0.540
& 0.681
& 67.78
& 0.609
& 0.431
& 0.774
& 0.865
& 0.376
& \textbf{3.11}
& 0.070 \\

\midrule
\multicolumn{18}{l}{\textit{Explicit 3D Memory}} \\

GEN3C
& 1,802
& 47.27
& 0.117
& 1.70
& 0.40
& 13.54
& 0.464
& 0.474
& 0.618
& 48.65
& 0.510
& 0.480
& \underline{0.813}
& 0.869
& 0.210
& 4.24
& 0.075 \\

Spatia
& 9,647
& 220.80
& 1.517
& 0.29
& 0.08
& 13.96
& 0.398
& 0.571
& 0.638
& 89.39
& 0.546
& 0.382
& 0.719
& 0.831
& 0.738
& 15.92
& 0.122 \\

SPMem
& 1,667
& 38.16
& 4.788
& 0.22
& 0.06
& 12.53
& 0.513
& 0.637
& 0.711
& 74.23
& 0.504
& 0.448
& 0.758
& 0.868
& 0.799
& 10.10
& \underline{0.056} \\

Lyra 2.0
& 1,273
& 92.27
& 4.231
& 11.36
& 1.69
& 13.60
& 0.382
& 0.493
& 0.618
& 52.26
& 0.651
& 0.481
& 0.801
& 0.882
& 0.348
& 4.79
& 0.074 \\

3DScenePrompt
& 149,880
& 6,292.00
& 5.635
& 0.24
& 0.06
& 13.63
& \textbf{0.528}
& 0.609
& 0.706
& 87.99
& 0.544
& 0.400
& 0.704
& 0.810
& 0.759
& 11.22
& 0.080 \\

NeoVerse
& 8,766
& 501.57
& 12.519
& 55.56
& 3.11
& 13.53
& \underline{0.525}
& 0.652
& 0.755
& 152.25
& 0.417
& 0.305
& 0.649
& 0.797
& 0.508
& 7.80
& 0.089 \\

\midrule

\textbf{\ours}
& \textbf{33}
& \textbf{11.88}
& \textbf{40.088}
& \textbf{460.73}
& \textbf{1,086.58}
& \textbf{16.00}
& 0.441
& \textbf{0.366}
& \textbf{0.547}
& \textbf{40.57}
& \underline{0.658}
& 0.497
& \textbf{0.820}
& \textbf{0.893}
& \textbf{0.142}
& \underline{3.41}
& 0.078 \\

\textbf{\ours Fast}
& \underline{70}
& \underline{25.55}
& \underline{37.814}
& \underline{323.25}
& \underline{823.71}
& \underline{14.59}
& 0.395
& \underline{0.430}
& \underline{0.569}
& \underline{43.05}
& \textbf{0.672}
& \textbf{0.515}
& 0.810
& \underline{0.889}
& \underline{0.200}
& 5.47
& 0.091 \\

\midrule

\multicolumn{18}{c}{\textit{RealEstate10K}} \\
\midrule

\multicolumn{18}{l}{\textit{Temporal Context}} \\

Yume 1.5
& --
& --
& --
& --
& --
& 15.22
& 0.652
& 0.634
& 0.698
& 48.01
& 0.618
& \underline{0.490}
& 0.796
& 0.881
& 0.818
& 10.19
& 0.048 \\

Matrix Game 3.0
& --
& --
& --
& --
& --
& \underline{17.95}
& 0.657
& 0.437
& 0.752
& 178.49
& 0.475
& 0.340
& 0.647
& 0.803
& 0.618
& 3.37
& 0.076 \\

\midrule
\multicolumn{18}{l}{\textit{Explicit 3D Memory}} \\

GEN3C
& 1,802
& 42.27
& 0.101
& 1.89
& 0.44
& 16.62
& 0.653
& 0.358
& 0.537
& 43.76
& 0.572
& 0.472
& 0.819
& 0.905
& \underline{0.112}
& \underline{0.46}
& 0.027 \\

Spatia
& 2,849
& 67.52
& 1.740
& 0.88
& 0.22
& 16.93
& 0.628
& 0.388
& 0.578
& 59.70
& 0.553
& 0.414
& 0.797
& 0.889
& 0.399
& 2.43
& 0.052 \\

SPMem
& 655
& \underline{14.99}
& 5.995
& 0.78
& 0.19
& 13.76
& 0.660
& 0.577
& \underline{0.531}
& 54.12
& 0.617
& 0.480
& 0.789
& 0.857
& 0.672
& 4.08
& \textbf{0.023} \\

Lyra 2.0
& 1,273
& 92.27
& 4.350
& 11.63
& 1.70
& 17.30
& 0.607
& \underline{0.313}
& 0.550
& \underline{37.36}
& \textbf{0.720}
& \textbf{0.512}
& \underline{0.824}
& 0.893
& \textbf{0.091}
& \textbf{0.20}
& \underline{0.025} \\

3DScenePrompt
& 149,880
& 6,292.00
& 5.642
& 0.28
& 0.07
& 15.46
& \textbf{0.700}
& 0.470
& 0.648
& 64.81
& 0.484
& 0.418
& 0.773
& 0.858
& 0.572
& 3.45
& 0.033 \\

NeoVerse
& 10,437
& 597.21
& 12.968
& 55.56
& 3.07
& 12.87
& 0.542
& 0.638
& 0.737
& 142.75
& 0.515
& 0.361
& 0.652
& 0.799
& 0.759
& 5.98
& 0.061 \\

\midrule

\textbf{\ours}
& \textbf{25}
& \textbf{8.91}
& \underline{38.881}
& \textbf{577.00}
& \underline{1,101.94}
& \textbf{18.36}
& \underline{0.668}
& \textbf{0.269}
& \textbf{0.518}
& \textbf{36.09}
& \underline{0.653}
& 0.454
& \textbf{0.841}
& \textbf{0.923}
& 0.216
& 1.23
& 0.029 \\

\textbf{\ours Fast}
& \underline{42}
& 15.15
& \textbf{42.823}
& \underline{516.10}
& \textbf{1,209.89}
& 16.19
& 0.609
& 0.442
& 0.579
& 49.94
& 0.590
& 0.445
& 0.799
& \underline{0.907}
& 0.474
& 1.02
& 0.030 \\

\bottomrule
\end{tabular}
}
\vspace{-10pt}
\end{table*}

\begin{figure}[t]
    \centering
   \vspace{-10pt}
   \includegraphics[width=\linewidth]{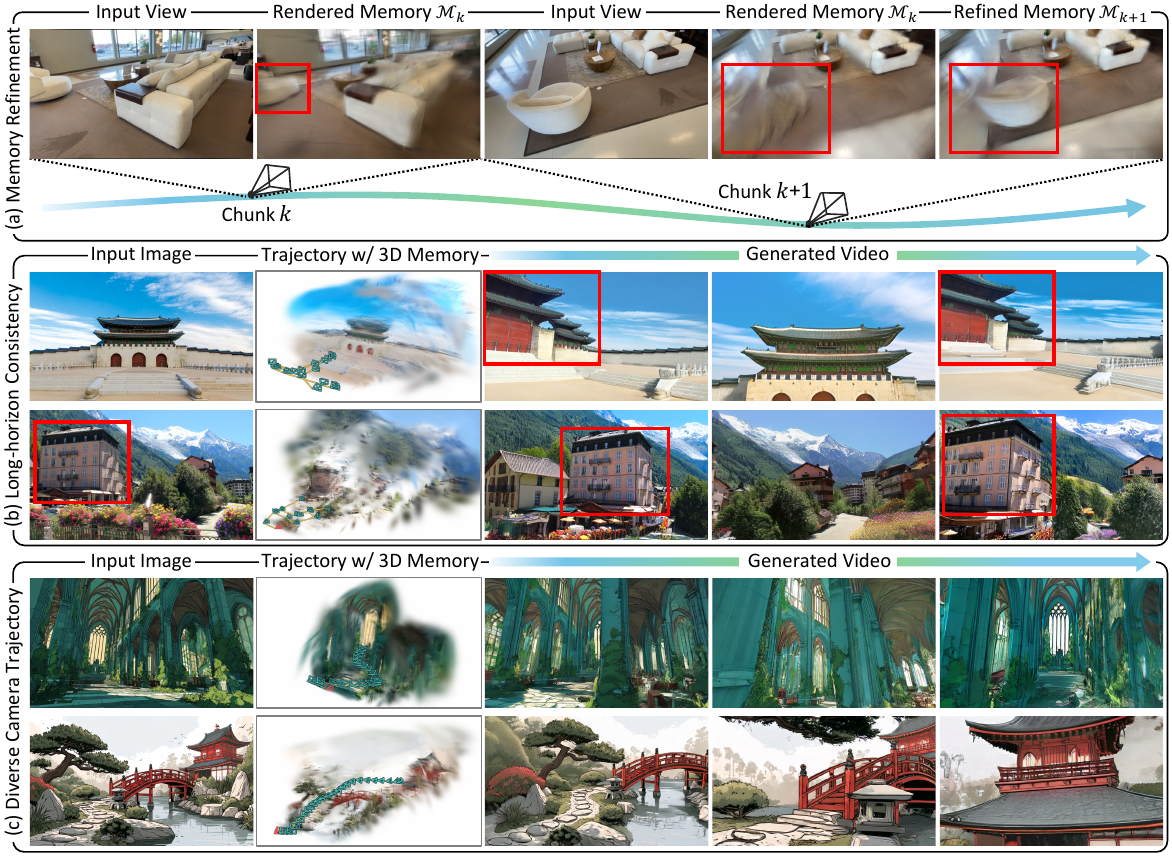}
    \vspace{-15pt}
    \caption{
    \textbf{Qualitative results of \ours.}
    (a) \oursmemupdater corrects geometric errors in the 3D memory through residual refinement using video chunks generated by \oursvid.
    (b) \ours maintains scene consistency over long-horizon generation through its persistent 3D memory.
    (c) Rendering the memory's VAE features from target camera views provides spatially aligned conditioning, enabling consistent video generation along diverse camera trajectories.
    }
    \label{fig:qual_video_ours}
    \vspace{-10pt}
\end{figure}

\subsection{Quantitative Results}

Tab.~\ref{tab:main_eval_video_l3g_world} compares \ours and \ours Fast with the baselines in terms of long-horizon self-consistency, memory efficiency, generation quality, camera pose accuracy, and 3D consistency.

\paragrapht{Memory Efficiency}
Compared with NeoVerse, a Gaussian-based method, \ours uses 266--417$\times$ fewer primitives and 42--67$\times$ less memory across the two datasets.
Its compact representation enables 8.3--10.4$\times$ faster RGB rendering, while directly rendering the stored VAE latent features increases VAE-conditioning throughput by 349--359$\times$. The advantage also holds against point-cloud-based memories.
Compared with Lyra 2.0, a particularly efficient point-cloud baseline, \ours performs memory reconstruction 8.9--9.5$\times$ faster while using 7.8--10.4$\times$ less memory, and achieves 643--648$\times$ higher VAE-rendering throughput. These gains directly reflect our two design choices: instead of accumulating dense primitives, \oursmemupdater adds Gaussians only for previously uncovered or high-error regions and updates existing ones; moreover, storing VAE features in each Gaussian eliminates the cost of VAE re-encoding required for conditioning the DiT.

\paragrapht{Self-Consistency}
On DL3DV, the two \ours variants achieve the best and second-best PSNR and LPIPS, with \ours obtaining the best scores. On RealEstate10K, \ours further achieves the best PSNR and LPIPS, and the second-best SSIM. These results support our motivation that a persistent, spatially anchored scene memory can preserve previously observed content when the camera revisits a region after long-horizon exploration.

\paragrapht{Generation Quality and 3D Consistency}
Importantly, these efficiency gains are accompanied by strong generation quality.
On DL3DV, \ours achieves the best LPIPS, FID, SC, and BC, while \ours Fast obtains the best IQ and AQ. On RealEstate10K, \ours achieves the best LPIPS, FID, SC, and BC, together with the second-best IQ. For camera control, \ours achieves the lowest translation error and second-lowest rotation error on DL3DV. While other baselines achieve lower pose and MEt3R errors on RealEstate10K, \ours maintains strong generation quality and long-horizon self-consistency.

\begin{figure}[t]
    \centering
   \vspace{-10pt}
   \includegraphics[width=\linewidth]{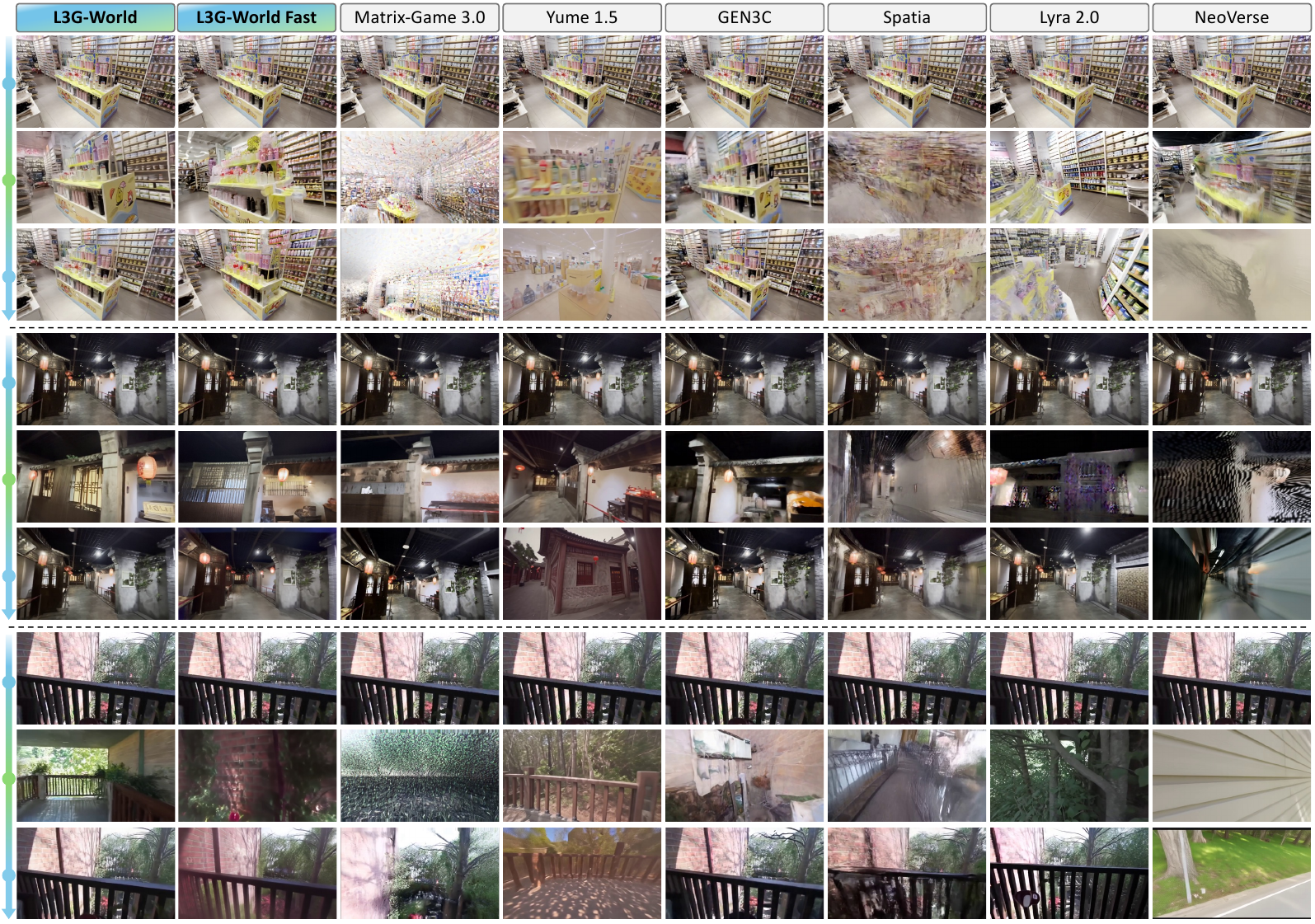}
    \vspace{-15pt}
    \caption{
    \textbf{Qualitative results of video generation on DL3DV and RealEstate10K.} For each scene, all methods start from the same input view (top), explore unseen regions (middle), and return to the initial camera location (bottom). \ours preserves scene structure and appearance upon loop closure, whereas the baselines exhibit drift, blur, or rendering artifacts.
    }
    \label{fig:qual_vieo_dl3dv_re10k}
\end{figure}
\subsection{Qualitative Results}
Fig.~\ref{fig:qual_video_ours} presents qualitative results of \ours.
As shown in (a), \oursmemupdater corrects geometric errors in the persistent 3D memory using observations from the next video chunk.
Specifically, rendering the memory $\mathcal M_k$ from chunk $k$ into the camera views of chunk $k{+}1$ reveals geometric errors (red bounding box).
Using the newly generated video chunk $k{+}1$, \oursmemupdater refines the existing Gaussians while spawning new ones in previously unobserved regions.
This adaptive memory update enables \ours to maintain a compact and efficient 3D memory while preserving scene consistency throughout minute-long autoregressive generation (b).
Furthermore, by rendering the memory's VAE features from target camera views, \ours provides spatially aligned conditioning to extrapolate the world from an input image along diverse camera trajectories (c).

Fig.~\ref{fig:qual_vieo_dl3dv_re10k} presents qualitative comparisons under long-horizon loop-closure trajectories. For each scene, all methods start from the same input view, explore previously unseen regions, and eventually return to the initial camera location; thus, the final frame directly reveals whether the generated world remains consistent after long-horizon exploration. Temporal-context-based methods, Matrix-Game 3.0 and Yume 1.5, often exhibit noticeable appearance or geometric drift when returning to previously observed regions. Among explicit 3D-memory approaches, Spatia and NeoVerse show substantial blur and rendering artifacts as generation progresses, while GEN3C and Lyra 2.0 produce comparatively stronger results but often struggle to extrapolate plausible geometry and appearance far beyond the single input view. In contrast, \ours preserves scene structure and appearance throughout long-horizon exploration and recovers a visually consistent scene upon loop closure. \ours Distill also retains this behavior while enabling few-step generation. Additional qualitative comparisons with all baselines are provided in Sec.~\ref{sec:appendix_qual}.

\subsection{Ablation Study}

\begin{table*}[t]
\centering
\caption{
\textbf{Ablation study of \oursmemupdater on DL3DV.}
We report RGB and VAE latent NVS quality and efficiency. Reconstruction FPS is computed from nine-frame update chunks.
The best and second-best results are shown in \textbf{bold} and \underline{underline}, respectively.
}
\label{tab:ablation_gs}
\vspace{3pt}
\resizebox{\textwidth}{!}{
\begin{tabular}{lcccccccc}

\toprule

Variant
& VAE MSE $\downarrow$
& VAE Cosine $\uparrow$
& \# Gaussians (K) $\downarrow$
& Recon. FPS $\uparrow$
& Render FPS (VAE) $\uparrow$
& PSNR $\uparrow$
& SSIM $\uparrow$
& LPIPS $\downarrow$
\\

\midrule

w/o Adaptive Spawning
& \underline{0.182}
& \underline{0.776}
& 75
& 39.08
& 742.4
& 16.26
& \textbf{0.409}
& \underline{0.535}
\\

w/o Refinement
& 0.213
& 0.743
& \underline{45}
& \textbf{43.82}
& 937.5
& 15.52
& 0.382
& 0.571
\\

w/o Global Self-Attn.
& 0.184
& 0.775
& \textbf{41}
& \underline{40.99}
& 938.5
& \underline{16.36}
& 0.397
& \underline{0.535}
\\

w/o RGB Loss
& 0.251
& 0.696
& 74
& 39.55
& \textbf{1199.0}
& 10.26
& 0.272
& 0.751
\\

w/o VAE Lifting
& 0.570
& 0.467
& --
& --
& 2.97
& --
& --
& --
\\

\midrule

\textbf{\oursmemupdater (Full)}
& \textbf{0.177}
& \textbf{0.782}
& \textbf{41}
& 40.03
& \underline{943.6}
& \textbf{16.52}
& \underline{0.406}
& \textbf{0.534}
\\

\bottomrule

\end{tabular}
}
\vspace{-10pt}
\end{table*}
\paragrapht{\oursmemupdater}
Tab.~\ref{tab:ablation_gs} ablates the key components of \oursmemupdater on novel view synthesis (NVS) on DL3DV, with all variants trained using a batch size of 2. We provide details of evaluation of \oursmemupdater in Sec.~\ref{sec:appendix_mem_eval}.
Without adaptive spawning, by applying K-means to all geometry features, the model uses $\sim$1.8$\times$ more Gaussians while degrading both RGB and VAE NVS, showing that selective spawning reduces redundant primitives.
Removing the refinement stream causes a meaningful quality drop, demonstrating the importance of correcting existing Gaussians through residual updates.
Removing global self-attention between the spawn and refine streams further degrades performance, highlighting the benefit of coordinating newly spawned and existing scene content.
We also ablate RGB supervision by setting $\lambda_{\mathrm{lpips}}^k{=}\lambda_{\mathrm{rgb}}^k{=}0$ for all $k$.
Although \oursvid is conditioned on directly rendered VAE features, RGB supervision remains important for learning accurate 3D geometry.
Finally, encoding RGB renderings with the VAE instead of lifting VAE features not only substantially slows down rendering but also degrades the rendered latent quality.
Overall, the full model achieves the strongest reconstruction quality across most RGB and VAE metrics while retaining a compact Gaussian memory and comparable reconstruction throughput.

\begin{figure}[t]
    \centering
   \vspace{-10pt}
   \includegraphics[width=\linewidth]{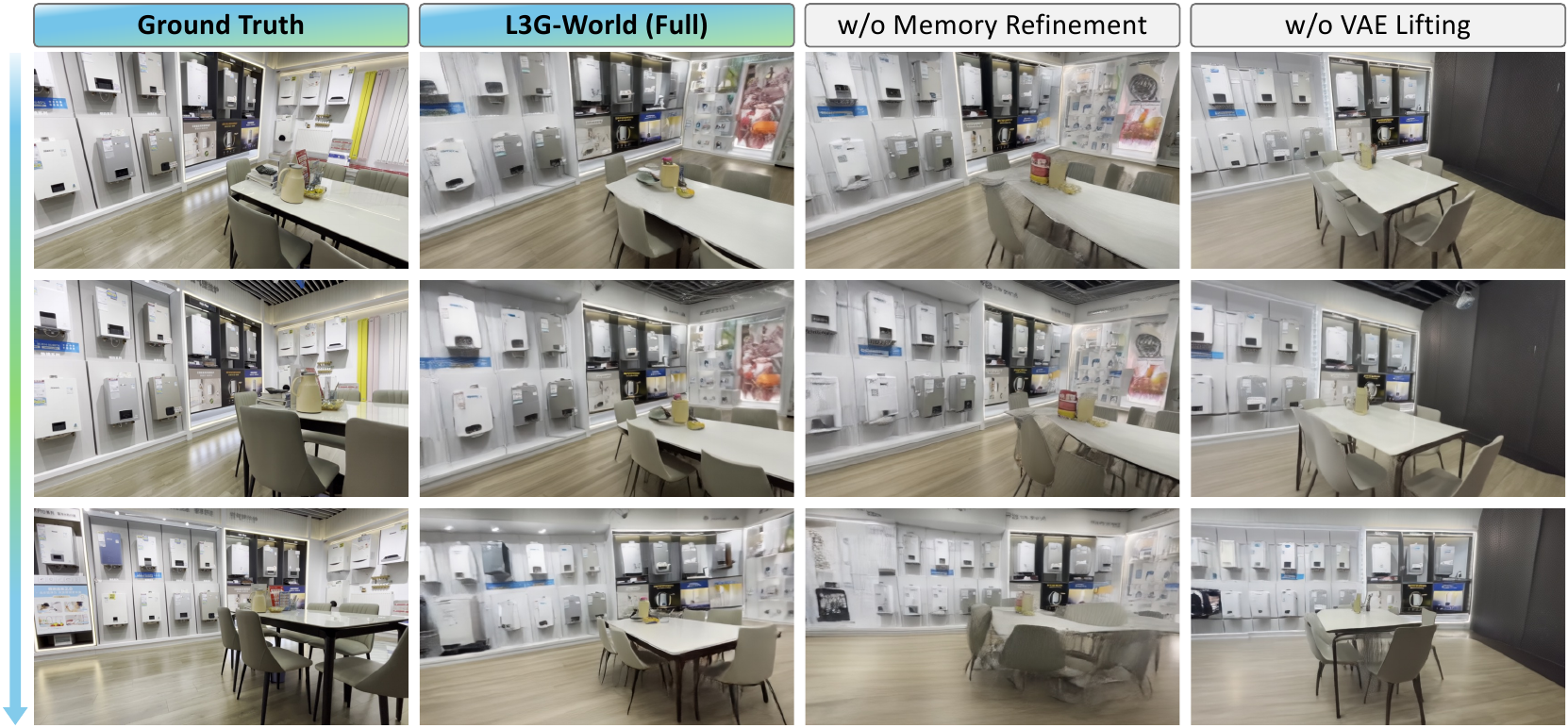}
    \vspace{-20pt}
    \caption{
    \textbf{Qualitative results of ablation study of \ours on DL3DV.} Memory refinement can not fix geometric error in 3D memory, leading blurry video due to memory rendering artifacts.
    }
    \vspace{-5pt}
    \label{fig:qual_video_abl}
\end{figure}
\begin{table*}[t]
\centering
\caption{\textbf{Ablation study of \ours on DL3DV}. The best and second-best results are shown in \textbf{bold} and \underline{underline}, respectively.
}
\vspace{3pt}
\label{tab:abl_dit}
\resizebox{\textwidth}{!}{%
\begin{tabular}{lccc|cccccc|cc|c}
\toprule
& \multicolumn{3}{c}{\textbf{Self-Consistency}}
& \multicolumn{6}{c}{\textbf{Generation Quality}}
& \multicolumn{2}{c}{\textbf{Pose Error}}
& \multicolumn{1}{c}{\textbf{3D Consistency}} \\
\cmidrule(lr){2-4}
\cmidrule(lr){5-10}
\cmidrule(lr){11-12}
\cmidrule(lr){13-13}
Method & PSNR $\uparrow$ & SSIM $\uparrow$ & LPIPS $\downarrow$ & LPIPS $\downarrow$ & FID $\downarrow$ & IQ $\uparrow$ & AQ $\uparrow$ & SC $\uparrow$ & BC $\uparrow$ & TransErr $\downarrow$ & RotErr ($^\circ$) $\downarrow$ & MEt3R $\downarrow$ \\
\midrule

w/o Memory Refinement
& 14.82
& 0.427
& 0.464
& \underline{0.592}
& \underline{50.63}
& \underline{0.605}
& 0.475
& 0.800
& 0.880
& \underline{0.314}
& 7.85
& 0.091 \\

w/o VAE Lifting
& \textbf{15.41}
& \textbf{0.480}
& \underline{0.461}
& 0.611
& 51.84
& 0.570
& \underline{0.478}
& \underline{0.808}
& \underline{0.885}
& 0.407
& \textbf{5.21}
& \textbf{0.075} \\

\textbf{\ours (Full)}
& \underline{15.11}
& \underline{0.431}
& \textbf{0.435}
& \textbf{0.583}
& \textbf{48.09}
& \textbf{0.619}
& \textbf{0.486}
& \textbf{0.814}
& \textbf{0.888}
& \textbf{0.278}
& \underline{5.22}
& \underline{0.081} \\

\bottomrule
\end{tabular}}
\vspace{-10pt}
\end{table*}

\paragrapht{\ours}
We further evaluate whether adaptively refining \oursmem, rather than simply accumulating primitives, improves long-horizon world generation.
We disable residual refinement in \oursmemupdater, such that 3D Gaussians remain fixed after being spawned and the memory evolves only through accumulation.
As shown in Tab.~\ref{tab:abl_dit}, removing memory refinement degrades all evaluation metrics, particularly self-consistency, pose accuracy, and 3D consistency.
This demonstrates that a persistent scene memory should not only preserve previously generated content but also progressively correct reconstruction error as the scene is revisited.
By continually refining the persistent 3D memory, \ours mitigates errors propagated over autoregressive generation and provides more reliable spatial conditioning for consistent and controllable world generation.

We also ablate direct VAE latent rendering by replacing it with RGB rendering followed by VAE encoding.
Direct latent rendering eliminates the additional encoding overhead and improves generation quality, highlighting its effectiveness for memory-conditioned video generation.


\section{Conclusion}

We presented \ours, a compact 3D Gaussian memory for real-time, spatially consistent world generation.
Instead of accumulating dense primitives from every generated chunk, our feed-forward memory updater spawns new Gaussians only in regions insufficiently covered by the memory and refines existing ones in previously reconstructed regions, keeping the memory compact and fast to render. Each Gaussian further carries a VAE-space feature, so the memory can be rendered directly into the latent space of the DiT without VAE re-encoding, which makes memory conditioning efficient enough for real-time generation. Experiments show that \ours achieves state-of-the-art generation quality in terms of long-horizon spatial consistency, while substantially reducing the cost of memory reconstruction, rendering, and storage. We believe that compact, latent-space 3D memories offer a practical foundation for scalable interactive world models.

\clearpage

\bibliography{iclr2027_conference}
\bibliographystyle{iclr2027_conference}

\newpage

\appendix

\section{Preliminaries}
\label{sec:prelim}

\subsection{Video Diffusion Models}
\label{sec:prelim_diffusion}
\paragrapht{Flow Map Formulation}
Video diffusion models~\citep{yang2025cogvideox, wan2025wan, ali2025world, agarwal2026cosmos, hacohen2026ltx} estimate the latent space of pretrained 3D VAE from pure Gaussian noise. Given video $x\sim p_{\mathrm{data}}$, the 3D VAE encoder $\mathcal{E}$ and decoder $\mathcal{D}$ yield $z_0=\mathcal{E}(x)$ and $\hat{x}=\mathcal{D}(z_0)$ where $z_0$ is the clean latent of video sample and $\hat{x}$ is decoded video. The latent state follows the probability-flow ODE $\dfrac{\mathrm{d}z_t}{\mathrm{d}t} = \mathbf{v}(z_t,t)$, whose flow map~\citep{sabour2026align} transports states from time $t$ to $r$:
\begin{equation}
\begin{aligned}
    \Phi_{r\leftarrow t}(z_t)
    &= z_t-\int_r^t \mathbf{v}(z_\tau,\tau)\,\mathrm{d}\tau \\
    &= z_t-(t-r)\mathbf{u}(z_t,r,t),
\end{aligned}
\end{equation}
where $0\leq r<t\leq1$ and $u$ denotes the average velocity. In the infinitesimal limit $r\to t$, $u(z_t,r,t)\to v(z_t,t)$, recovering the instantaneous velocity modeled by flow matching. For training samples $z_t=(1-t)z_0+t\epsilon$, with $\epsilon\sim\mathcal{N}(0,I)$ and $t\sim\mathcal{U}(0,1)$, a velocity predictor $v_\theta$ is trained using
\begin{equation}
    \mathcal{L}_{\mathrm{FM}}
    = \mathbb{E}_{x,\epsilon,t}
    \left[
        \left\|\mathbf{v}_\theta(z_t,t)
        -(\epsilon-z_0)\right\|_2^2
    \right].
    \label{eq:flow_matching}
\end{equation}

\paragrapht{On-Policy Distillation}
On-policy distillation~\citep{gu2026anyflow} alleviates the train--test discrepancy by optimizing the student on samples generated from its own inference trajectory under the guidance of a pretrained teacher. Given Gaussian noise $z$, the student first generates $\hat{z}_0=f_\theta(z)$ through its inference process. Following DMD-style reverse-KL optimization, the generated sample is re-noised at $t\in[0,T]$ as $z_t=(1-t)\hat{z}_0+t\epsilon$, where $\epsilon\sim\mathcal{N}(0,I)$, and the student is optimized using the discrepancy between the real and generated score functions:
\begin{equation}
\nabla_\theta \mathcal{L}_{\mathrm{DMD}}
=
-\mathbb{E}_{t,z}
\left[
\left(
s_{\mathrm{real}}(z_t,t)
-
s_{\mathrm{fake}}(z_t,t)
\right)
\frac{\partial f_\theta(z)}{\partial\theta}
\right].
\label{eq:dmd}
\end{equation}

\subsection{3D Gaussian Splatting}
\label{sec:prelim_gaussian}
3D Gaussian Splatting~\citep{kerbl3Dgaussians} represents a scene as a collection of anisotropic 3D Gaussians $\mathcal{G}=\{(\mu_i,\Sigma_i,\alpha_i,c_i)\}_{i=1}^{N}$, where $\mu_i\in\mathbb{R}^3$ denotes the Gaussian center, $\Sigma_i\in\mathbb{R}^{3\times3}$ its covariance, $\alpha_i$ its opacity, and $c_i$ its view-dependent color. The covariance is parameterized by a rotation and scale as $\Sigma_i=R_iS_iS_i^\top R_i^\top$. Given a camera viewpoint, the 3D Gaussians are projected onto the image plane and rendered via depth-ordered alpha compositing. The resulting pixel color is computed as
\begin{equation}
    C
    =
    \sum_{i=1}^{N}
    c_i \alpha_i
    \prod_{j=1}^{i-1}(1-\alpha_j),
\end{equation}
where $c_i$ and $\alpha_i$ denote the color and effective opacity of the $i$-th projected Gaussian at the corresponding pixel.

\section{Implementation Details}

\begin{table}[h]
\centering
\fontsize{8}{10}\selectfont
\setlength{\tabcolsep}{4pt}
\caption{
\textbf{Training configuration} of \oursmemupdater, \oursvid, and \oursvid Distilled.
For \oursvid Distilled, values separated by ``/'' denote the flow-map and
on-policy distillation stages, respectively.
}
\vspace{10pt}
\begin{tabular}{lccc}
\toprule
\textbf{Component}
& \oursmemupdater
& \oursvid
& \oursvid Fast \\
\midrule
Optimizer
& AdamW
& AdamW
& AdamW \\

Learning rate
& $1 \times 10^{-4}$
& $1 \times 10^{-4}$
& $5 \times 10^{-5}$ / $2 \times 10^{-6}$ \\

Optimizer betas
& $(0.9, 0.95)$
& --
& $(0.9, 0.95)$ / $(0, 0.999)$ \\

Weight decay
& 0.05
& 0.01
& - \\

Batch size
& 8
& 16
& 16 / 8 \\

EMA decay
& --
& --
& 0.999 / 0.99 \\

Gradient clipping
& 1.0
& 1.0
& -- \\

Training steps
& 94K
& 3.3K
& 1.0K / 4.6K \\

Total parameters
& 1.7B
& 2B
& 2B \\

Trainable parameters
& 1.6B
& 75.8M
& 210.3M \\

RGB Resolution ($H\times W$)
& $210 \times 364$
& $480 \times 832$
& $480 \times 832$ \\

VAE latent resolution ($H\times W$)
& $60 \times 104$
& $60 \times 104$
& $60 \times 104$ \\
\bottomrule
\end{tabular}
\label{tab:appendix_train_config}
\end{table}

\subsection{Gaussian Memory Updater}
\label{sec:appendix_impl_gs}

\paragrapht{Architecture}
At step $k$, the updater \oursmemupdater consumes the current window $\mathcal W_k=\{\mathbf I_k,\mathbf C_k,\mathbf z_k\}$ and the persistent memory $\mathcal M_{k-1}=\{\mathbf G_{k-1},\mathbf T_{k-1}\}$.
We feed the frames and cameras to the geometry backbone~\citep{da3} and extract $1{,}536$-dimensional geometry features $\mathbf F_k=\{\mathbf F^l_k\}_{l=1}^{L}$ from layers $39$, $29$, and $19$ ($L=3$); the cross-attention layers attend to geometric tokens in this order. The backbone patch size is $14$. In parallel, two convolutional encoders map the VAE latents $\mathbf z_k$ to $128$-dimensional VAE tokens.

K-means on the geometry feature $\mathbf F^1_k$ selects the indices that initialize the spawn queries $\mathbf Q_k$ and that gather the aligned queries from the remaining geometry layers, the RGB grid, and the VAE grid. The selected RGB and VAE queries additionally cross-attend to the complete RGB and VAE token sets.

The spawning stream and refinement stream each contain $L{=}3$ blocks of cross-attention, self-attention, and global self-attention followed by a feed-forward network. The spawn Gaussian head decode each resulting memory token $\bar{\textbf{T}}^\text{sp}_k$ into 32 Gaussians, following~\citep{veicht2026zipsplat}. The refine Gaussian head decode each retrieved memory token $\bar{\textbf{T}}^\text{rf}_k$ into residual parameters of 32 Gaussians. 
Finally, the new memory tokens $\bar{\textbf{T}}^\text{sp}_k$ are cached in the persistent memory set $\mathcal M_k$, while corresponding memory tokens and Gaussians are updated by $\bar{\textbf{T}}^\text{rf}_k$.

\paragrapht{Spawn Mask}
We compute the spawn mask $\mathbf S_k$ of Sec.~\ref{sec:method_gaussian} on the patch grid of $\mathrm F_k^1$.
For each spatial patch grid $p\in\{1, \ldots, Vhw\}$, where $h$ and $w$ are the spatial resolution of $\mathrm F_k^1$, we average the accumulated opacity and the per-pixel RGB mean squared error (MSE) over its $14\times14$ pixels to obtain downsampled mask as:
\begin{equation}
    \mathbf S_{k}(p)
    =
    \mathbbm{1}\!\left(
        E_{k}(p) > \tau_E
        \;\lor\;
        A_{k}(p) < \tau_\alpha
    \right),
\end{equation}
where $E_{k}(p)$ and $A_{k}(p)$ is the MSE and accumulated alpha at grid $p$, respectively.
Low-opacity patches thus propose Gaussians for uncovered regions, whereas high-error patches propose additional Gaussians in regions that are covered but poorly reproduced by the current memory.
The number of spawning queries $N_{\mathrm q}$, which serves as the number of clusters in K-means, is set in proportion to the fraction of masked patches in the current window, ranging linearly from $N_{\min}$ when no patch is masked to $N_{\max}$ when all patches are masked. Every query produces $32$ Gaussians, and at $k=1$, where the memory is empty, we use $N_{\max}$.

\paragrapht{Gaussian Retrieval}
At $k>1$, we rasterize $\mathbf G_{k-1}$ once from the current context cameras to obtain the rendered depth and projected footprint of each Gaussian.
For each Gaussian $\mathcal G_n$ projected into view $i$, we probe pixels at its projected center and along the principal axes of its projected ellipse.
A Gaussian is considered visible if at least one valid probe $u \in \{1, \ldots, VHW\}$ lies inside the image and satisfies
\begin{equation}
    d_n
    \leq
    \hat{\mathbf D}(u) + \tau_d,
    \qquad
    \alpha_n^\text{local}(u) \geq \tau_{\alpha^\text{local}},
\end{equation}
where $d_{n}$ is the projected depth of Gaussian $\mathcal G_n$, $\hat{\mathbf D}(u)$ is the depth rendered from the current Gaussian memory at pixel $u$, $\alpha_n^\text{local}(u)$ is the contribution of Gaussian $G_n$ at pixel $u$, and $\tau_d$ and $\tau_{\alpha^\text{local}}$ denote the depth and opacity thresholds, respectively.
The depth criterion removes Gaussians lying sufficiently behind the currently rendered surface, while the opacity criterion filters out Gaussians with negligible influence on rendering.
We use $\tau_d=0.05$ and $\tau_{\alpha^\text{local}}=1/255$ by default.
We retrieve all memory tokens associated with at least one visible Gaussian and pass them to the refinement stream.

\paragrapht{Residual Memory Update}
For each retrieved Gaussian $\mathcal{G}_n=({\mu}_n,{\Sigma}_n,\alpha_n,{c}_n,{z}_n) \in \mathbf{G}_{k-1}$, we perform a residual update of its parameters. We omit the Gaussian index $n$ and window index $k$ for clarity. The Gaussian covariance ${\Sigma}$ is parameterized by a scale ${s}$ and a quaternion ${q}$ as
\begin{equation}
    {\Sigma}
    =
    {R}({q})
    \operatorname{diag}({s}^2)
    {R}({q})^\top.
\end{equation}
Given the refine tokens $\bar{\mathbf{T}}^{\mathrm{rf}}_k$, the refine head predicts per-Gaussian residuals
$(\delta_\mu,\delta_s,\delta_q,\delta_\alpha,\delta_c,\delta_z)$
for the retrieved Gaussians, along with two independent gates, $g$ and $g_z$, for the Gaussian parameters and VAE latent feature, respectively.
We update the center, scale, quaternion, opacity, and color using residuals gated by $g$, and the VAE latent feature using the residual gated by $g_z$.
The covariance is then reconstructed from the updated scale and quaternion.

Finally, we update the previous memory
$\mathcal{M}_{k-1}=(\mathbf{G}_{k-1},\mathbf{T}_{k-1})$
by replacing the retrieved memory tokens with $\bar{\mathbf{T}}^{\mathrm{rf}}_k$ and applying the predicted residual updates only to the retrieved Gaussians, while leaving the remaining tokens and Gaussians unchanged.

\paragrapht{VAE Feature Lifting}
Using a simple convolutional network, we embed $16$-channel VAE latents $\mathbf z_k$ into $128$-dimensional tokens, resulting in features spatially aligned with the geometry features.
The K-means representative indices from spawning query tokens initialize the VAE queries for newly spawned tokens. A single cross-attention layer lets the VAE queries attend to all current VAE tokens. The spawn Gaussian head $\operatorname{Head}_{\text{sp}}(\cdot)$ therefore receives a concatenated token from spawn token and VAE token. 
For each token, the head predicts both the $32$ Gaussians and a 16-dimensional VAE latent feature for each Gaussian. VAE features are rendered using the same geometry, opacity, and compositing as RGB. We detach Gaussian geometry in this feature rasterizer, so the VAE objective supervises the lifted features and shared token representation without moving Gaussian centers through the VAE render.

\paragrapht{Training}
We train \oursmemupdater on $K{=}5$ consecutive video chunks, without frame overlap, each containing $V \in \{1,\ \ldots,\ 9\}$ input images with cameras and VAE latents. Ground-truth cameras are provided to both the geometry backbone and renderer. We place every rollout in a single coordinate system whose origin and orientation are defined by the first context camera of chunk $1$. A single scale, equal to the reciprocal of the maximum pairwise camera-center distance over the complete clip, is then applied to all camera translations and depths in every chunk. For depth supervision, we use depth maps predicted by Depth Anything 3~\citep{da3}.
We use $\tau_E=0.05$, $\tau_\alpha=0.25$, $N_\text{min}=1,024$, and $N_\text{min}=10,240$, by default during training and inference.

We use frozen VAE latents~\citep{nvidia2025cosmos2.5} as input, where every 4-th RGB frame of each VAE latent is used for input video frames. 
Training follows the online recurrence used at inference. After each chunk, to ensure temporal consistency, the keys and values in cross-attention and local self-attention layers of the spawning stream are cached and remain attached within the sampled rollout. At chunk $k$, the updated global memory is rendered into the target views of chunks $1,\ldots,k$.

The reconstruction losses are defined as:
\begin{equation}
    \label{eq:memory_loss}
    \begin{aligned}
        \mathcal{L}_{\mathrm{rgb}}
        &=
            \operatorname{mean}(\|\hat{\mathbf{I}}-\mathbf{I}^{\mathrm{tgt}}\|_1), \\
        \mathcal{L}_{\mathrm{lpips}}
        &=
        \operatorname{LPIPS}(
            \hat{\mathbf{I}}, \mathbf{I}^{\mathrm{tgt}}), \\
        \mathcal{L}_{\mathrm{depth}}
        &=
            \operatorname{mean}(\|\hat{\mathbf{D}}-\mathbf{D}^{\mathrm{tgt}}\|_1), \\
        \mathcal{L}_{\mathrm{vae}}
        &=
            \operatorname{mean}(\|\hat{\mathbf{z}}-\mathbf{z}^{\mathrm{tgt}}\|_2^2),
    \end{aligned}
\end{equation}
where $\hat{\mathbf I}$, $\hat{\mathbf D}$, and $\hat{\mathbf z}$ are the RGB, depth, and VAE feature renderings of the memory, and $\mathbf I^{\mathrm{tgt}}$, $\mathbf D^{\mathrm{tgt}}$, and $\mathbf z^{\mathrm{tgt}}$ are the corresponding target images, Depth Anything 3 depth maps, and VAE latents. Here, the mean is taken over the corresponding pixels or latent elements across target views, and LPIPS is averaged over target images. Following TokenGS~\citep{tokengs2026}, we additionally use a visibility loss $\mathcal{L}_{\mathrm{vis}}$ that encourages each Gaussian to lie within at least one observed input-view frustum. We omit the window index from the loss terms for simplicity.
At each rollout step, we evaluate the weighted objective in Eq.~\ref{eq:reconloss} and minimize its average over the $K$ chunks. Although previous memory states are detached, supervision from earlier target views still constrains the current spawning and refinement outputs, encouraging consistency with previously reconstructed content.

We initialize our memory updater \oursmemupdater from ZipSplat~\citep{veicht2026zipsplat}, while the new refine stream, global self-attention, and convolutional network for VAE latent are trained from scratch. We train \oursmemupdater end-to-end, including the geometric foundation model. When widening the Gaussian head for jointly decoding Gaussian primitives and VAE features, we copy all pretrained weights exactly and zero-initialize its new VAE-input columns.

\subsection{Video Diffusion Model Details}
\label{sec:appendix_impl_diffusion}

\subsubsection{Teacher Model}
\paragrapht{Architecture}
We build the teacher model on Cosmos-Predict2.5-2B~\citep{ali2025world}.
The pretrained DiT backbone is kept frozen, while LoRA adapters, the UCPE~\citep{zhang2026unified} camera-conditioning branches, and the widened input projection are trained.
For memory conditioning, the noisy video latent is channel-concatenated with the VAE feature rendered from the current \oursmem, together with its validity mask and conditioning indicator.
We additionally retrieve up to three previously observed latents according to camera-pose proximity and provide them as clean reference tokens.

\paragrapht{Training}
We train the teacher using rectified flow matching on DL3DV~\citep{ling2024dl3dv} and RealEstate10K~\citep{zhou2018stereomagnificationlearningview}.
The anchor frame is provided as a clean image latent and excluded from the denoising loss, while the remaining target latents are jointly denoised with full bidirectional attention.
During training, we randomly drop text and reconstruction conditioning and vary the number of retrieved reference latents to improve robustness to different conditioning configurations.
The teacher is fine-tuned with rank-16 LoRA adapters~\citep{hu2022lora} together with the UCPE modules and input projection.

\paragrapht{Inference}
At inference, the teacher generates one 81-frame clip at a time using 35 denoising steps with classifier-free guidance~\cite{ho2022classifier}.
After each generated clip, the reconstructed scene memory is updated from the generated observations and rendered under the cameras of the next clip.
Reference latents are retrieved only from the generated history, and the final generated frame is used to initialize the following clip.

\subsubsection{Student Model}
\paragrapht{Architecture}
The student retains the same backbone, reconstruction conditioning, reference conditioning, and UCPE camera modules~\citep{zhang2026unified}  as the teacher, but converts the bidirectional DiT into a chunk-causal autoregressive model.
Each training clip is divided into short latent chunks, and the current chunk attends to compressed long-term history, retrieved reference latents, and the most recent full-resolution chunks.
Older history is spatially compressed, while the two most recent chunks remain at the original latent resolution.
At inference, the corresponding keys and values are cached independently for the standard self-attention and UCPE attention branches.

\paragrapht{Flow-Map Distillation}
We first initialize the student through flow-map distillation following AnyFlow~\citep{gu2026anyflow}.
Unlike the teacher, which predicts the instantaneous denoising velocity at a single noise level, the student is conditioned on both the source and destination noise levels and learns transitions between arbitrary points along the denoising trajectory.
Training combines standard diffusion, direct consistency, and intermediate flow-map transitions.
We initialize the student from the teacher and increase the LoRA rank to 128, while preserving the teacher initialization for the shared camera and reconstruction-conditioning modules.

\paragrapht{On-Policy Distillation and Inference}
We further train the student with on-policy distribution matching using samples generated from its own autoregressive rollouts.
A frozen teacher provides the real-data score, while a trainable copy of the teacher estimates the student distribution, and the flow-map objective~\citep{boffi2024flow} is retained during this stage for stable optimization.
At inference, the student generates three latent frames per autoregressive step using 4 denoising steps without classifier-free guidance.
The generated latents are continuously added to the reference pool and periodically used to update the \oursmem, which is then rendered to provide reconstruction guidance for subsequent chunks.

\subsection{Baselines}
\label{sec:appendix_baseline}

\paragrapht{Depth and Scale Alignment}
Our evaluation considers a single-image-to-video setting, where the explicit 3D-memory baselines~\citep{ren2025gen3c,zhao2026spatia,wu2026spmem,shen2026lyra,lee20263d,yang2026neoverse} must initialize their scene representation from a single input image. This introduces a scale ambiguity between the monocularly reconstructed 3D scene and the metric camera trajectory used for evaluation, which can lead to incorrectly rendered target views even when the reconstructed geometry is otherwise accurate. To ensure a consistent metric scale across methods, we use Depth Anything 3 GIANT-1.1-NESTED~\citep{da3} as the depth estimator for all explicit 3D-memory baselines. The predicted metric depth is used to place the reconstructed 3D representation at the same scale as the evaluation camera trajectory. For Gaussian-based methods, NeoVerse~\citep{yang2026neoverse} and \ours, we additionally align the depth rendered from the initially predicted Gaussians with the metric depth estimated from the input image.

\paragrapht{Baseline Implementations}
The official implementations of SPMem~\citep{wu2026spmem} and 3DScenePrompt~\citep{lee20263d} initialize their first generation chunk using multiple input views or video frames, from which they reconstruct a 3D point-cloud representation and render it under subsequent target cameras. Since our benchmark provides only a single input image, we modify their initialization stage to reconstruct the 3D representation from the first frame alone and render it to the target cameras of the first chunk. For subsequent autoregressive generation chunks, we retain the behavior of the official implementations.

NeoVerse~\citep{yang2026neoverse} is designed for single-image world generation, but its released implementation does not provide the complete autoregressive generation pipeline required by our long-horizon evaluation. We therefore build an autoregressive inference pipeline on top of the official implementation, repeatedly reconstructing and updating its Gaussian representation from generated observations and rendering the resulting memory under the target cameras for subsequent generation.

Yume 1.5~\citep{mao2026yume1} and Matrix-Game 3.0~\citep{wang2026matrix} are action-conditioned world models that use discrete keyboard-style controls rather than continuous camera parameters. To evaluate them along the same camera trajectories as the other methods, we convert the provided camera motion into the corresponding sequence of discrete action inputs and use these actions for autoregressive generation.

\subsection{Evaluation}
\label{sec:appendix_eval}

\paragrapht{Datasets}
We evaluate \ours and all baselines on the test splits of DL3DV~\citep{ling2024dl3dv} and RealEstate10K~\citep{zhou2018stereomagnificationlearningview}. To explicitly evaluate long-horizon consistency under revisitation, we construct 481-frame loop-closure trajectories. Starting from the input view, the camera first follows an exploration trajectory and subsequently retraces the same trajectory in reverse, eventually returning to the initial camera pose.

\paragrapht{Self-Consistency}
The loop-closure trajectory provides pairs of generated frames corresponding to the same camera pose but generated at different points in the autoregressive sequence. We pair each frame on the outgoing trajectory with its corresponding frame on the returning trajectory and compute PSNR, SSIM, and LPIPS between them. We report the average over all corresponding frame pairs as our self-consistency metrics. This evaluation directly measures whether scene appearance and structure are preserved after long-horizon generation and subsequent loop closure.

\paragrapht{Camera Controllability}
Following~\citep{seo2026grounding}, we estimate the camera trajectory from each generated video using the feed-forward geometry estimator~\citep{da3}, and compare the estimated trajectory against the ground-truth camera trajectory used for generation. We report translation error (TransErr) after resolving the global scale ambiguity of the estimated trajectory, together with rotation error (RotErr) in degrees.

\paragrapht{Efficiency}
For explicit 3D-memory methods, we measure the number of stored 3D primitives and their memory footprint after long-horizon generation, together with memory reconstruction throughput in processed views per second. We further measure target-view rendering throughput in two settings: RGB rendering FPS and VAE-space conditioning FPS. For methods whose 3D memory stores only RGB appearance, VAE-space conditioning includes both target-view RGB rendering and subsequent VAE encoding. In contrast, \ours directly renders the stored VAE features into the target-view latent space without RGB rendering or re-encoding. Reconstruction FPS includes Gaussian retrieval, mask computation, the updater forward pass, and the residual update. All efficiency measurements in Tab.~\ref{tab:main_eval_video_l3g_world} are performed on a single NVIDIA A100 GPU.

\section{Qualitative Results}
\label{sec:appendix_qual}

\begin{figure}[t]
    \centering
   \vspace{-10pt}
   \includegraphics[width=\linewidth]{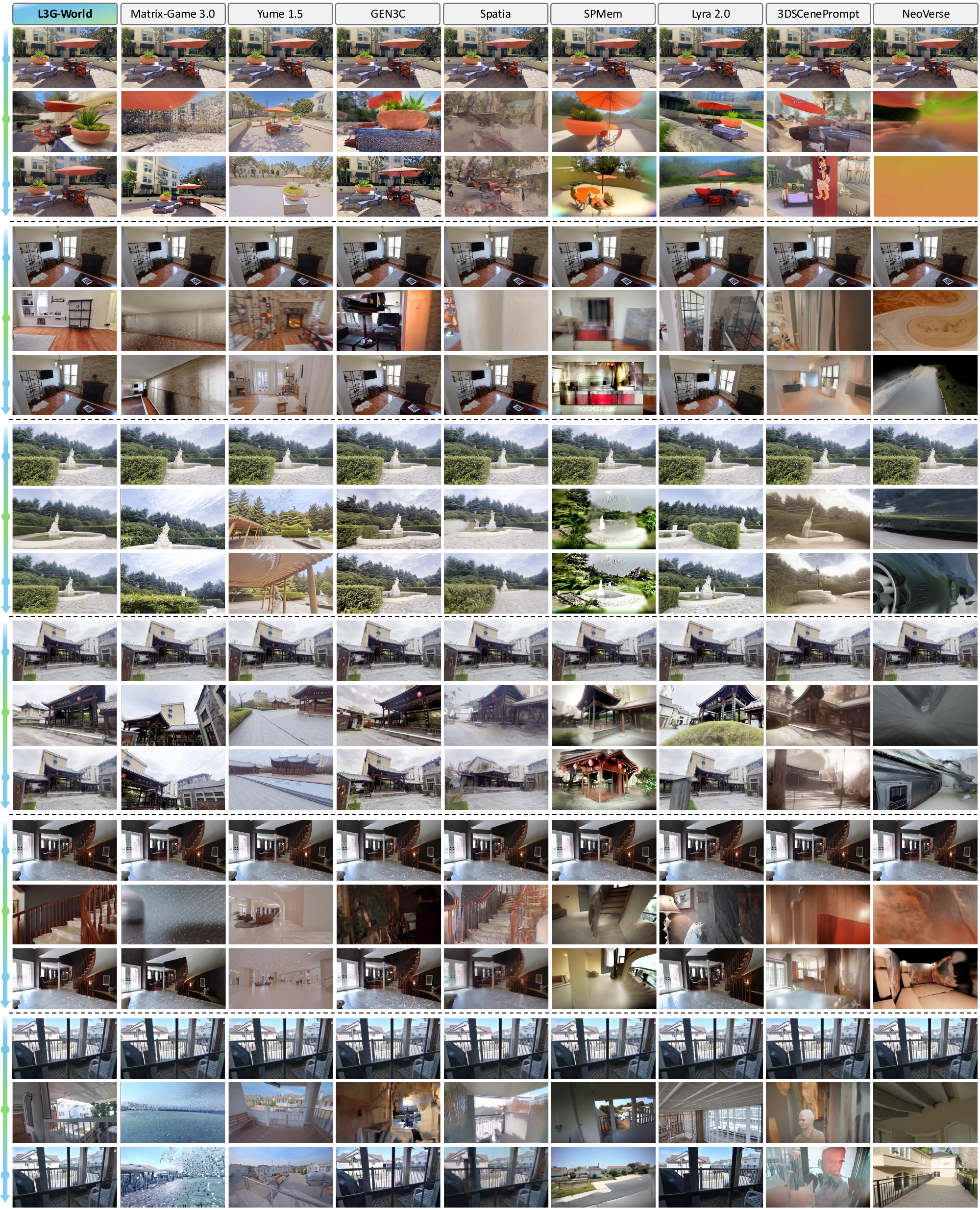}
    \vspace{-15pt}
    \caption{
    \textbf{Additional qualitative results of video generation on DL3DV and RealEstate10K.} For each scene, all methods start from the same input view (top), explore unseen regions (middle), and return to the initial camera location (bottom). \ours preserves scene structure and appearance upon loop closure, whereas the baselines exhibit drift, blur, or rendering artifacts.
    }
    \label{fig:qual_video_appendix}
\end{figure}

Fig.~\ref{fig:qual_video_appendix} extends the comparison in Fig.~\ref{fig:qual_vieo_dl3dv_re10k} to additional scenes from DL3DV and RealEstate10K under the same long-horizon loop-closure trajectories. Starting from the same input view, each method explores previously unseen regions and eventually returns to the initial camera location, so that the final frame directly reveals whether the generated world remains consistent after long-horizon exploration.
The baselines exhibit distinct failure modes. 
Temporal-context-based methods, Yume 1.5 and Matrix-Game 3.0, generally follow the camera motion during exploration, but fail to preserve spatial consistency upon return: the final frame often shows a layout or appearance different from the input view, and in some cases even a different viewpoint, indicating that camera control also degrades over the long trajectory.
Among explicit 3D memories, SPMem tends to hallucinate inconsistent content once the camera leaves the initially observed region, and NeoVerse progressively collapses into blurry textures over the long trajectory.
Among the baselines that return to the starting scene more reliably, GEN3C recovers the input view quite well, but struggles to extrapolate plausible content in regions unseen in the input view, producing distorted or incomplete frames during exploration. 
Lyra 2.0 returns reasonably close to the input view, but its final frame is still not fully spatially aligned with it, indicating drift from the original scene.
In contrast, \ours both generates plausible content in unseen regions and returns to a scene spatially aligned with the input view in both structure and appearance, as its latent 3D Gaussian memory anchors previously generated content while continually refining it throughout exploration.

\section{Comparison with Online 3DGS Methods}
\label{sec:appendix_mem_eval}

Existing online 3D Gaussian Splatting methods~\cite{cheng2026recosplat, chen2026of3gs} provide strong feed-forward reconstruction and could, in principle, serve as an external 3D memory for world generation. However, these methods are primarily designed for high-fidelity reconstruction and novel-view synthesis (NVS), whereas a memory for real-time world generation must satisfy a different operating regime: it must be continuously updated throughout autoregressive generation, remain compact as exploration proceeds, and render the conditioning signal for every subsequent video chunk with very low latency. In particular, a dense representation that is effective for reconstruction can introduce substantial memory and rendering overhead when repeatedly queried by a video diffusion model. Our goal is therefore not to outperform dedicated online 3DGS methods as a standalone reconstruction system, but to construct a sufficiently faithful 3D memory that can be efficiently updated and directly consumed by \oursvid.

To characterize this trade-off quantitatively, we evaluate \oursmem with online 3D Gaussian reconstruction models~\cite{cheng2026recosplat, chen2026of3gs} on DL3DV~\citep{ling2024dl3dv}, and additionally on RealEstate10K~\citep{zhou2018stereomagnificationlearningview} for out-of-domain generalization. We process each sequence online in streaming chunks of nine frames. After each chunk processing, we render the current 3D Gaussians to four target viewpoints and evaluate both RGB- and VAE-space NVS. We report PSNR, SSIM, and LPIPS for RGB NVS, and mean squared error (MSE) and cosine similarity between the rendered and ground-truth VAE latents. We additionally measure the number of stored primitives, memory footprint, reconstruction FPS, and VAE rendering throughput to evaluate the efficiency required for repeated memory updates and queries during world generation.

As shown in Tab.~\ref{tab:recon}, \oursmem is substantially more efficient than dedicated online 3DGS methods~\citep{cheng2026recosplat, chen2026of3gs}. Compared with ReCoSplat~\citep{cheng2026recosplat} at $480\times480$, \oursmem uses approximately $49$--$53\times$ fewer primitives and $13$--$14\times$ less memory across the two datasets, while updating the memory about $2.5$--$2.6\times$ faster. 
Moreover, since these methods store only RGB Gaussians, conditioning a latent video generator requires rendering RGB images and re-encoding them with the VAE, whereas \oursmem directly renders its stored VAE features, achieving $245$--$430\times$ higher VAE rendering throughput. 
Despite this compact representation, \oursmem retains meaningful reconstruction fidelity: at our target resolution of $480\times832$, it substantially improves both RGB- and VAE-space NVS over ReCoSplat on both datasets, and outperforms OF$^3$GS~\citep{chen2026of3gs} while operating at a higher resolution. ReCoSplat at the lower $480\times480$ resolution achieves higher fidelity, which is expected from a method specifically optimized for high-quality NVS. We therefore view these reconstruction metrics not as a standalone reconstruction benchmark, but as an objective measure that \oursmem preserves sufficiently accurate spatial and latent information while satisfying the much stricter memory and throughput requirements of real-time world generation.

\begin{table*}[t]
\centering
\caption{
\textbf{Online NVS evaluation.}
We evaluate online reconstruction on DL3DV and on RealEstate10K, reporting RGB and VAE latent novel view synthesis quality together with memory size, reconstruction FPS, and VAE rendering FPS. Reconstruction FPS is computed from nine-frame update chunks. The best and second-best results within each dataset are shown in \textbf{bold} and \underline{underline}, respectively.
}
\label{tab:recon}
\resizebox{\textwidth}{!}{
\begin{tabular}{llccccccccc}
\toprule

Method
& \shortstack{Resolution\\$(H\times W)$}
& \shortstack{VAE\\MSE $\downarrow$}
& \shortstack{VAE Cosine\\Sim. $\uparrow$}
& \shortstack{\# Primitives\\(K) $\downarrow$}
& \shortstack{Memory\\(MiB) $\downarrow$}
& \shortstack{Recon.\\FPS $\uparrow$}
& \shortstack{VAE Render\\FPS $\uparrow$}
& PSNR $\uparrow$
& SSIM $\uparrow$
& LPIPS $\downarrow$
\\
\midrule

\multicolumn{11}{c}{\textit{DL3DV}} \\
\midrule

ReCoSplat
& $480\!\times\!480$
& \textbf{0.118}
& \textbf{0.858}
& \underline{2,085}
& \underline{198.88}
& \underline{11.12}
& \underline{3.0}
& \textbf{21.53}
& \textbf{0.707}
& \textbf{0.318} \\

ReCoSplat
& $480\!\times\!832$
& 0.664
& 0.344
& 11,550
& 1101.48
& 2.15
& 2.7
& 12.41
& 0.438
& 0.711 \\

OF$^3$GS
& $405\!\times\!832$
& 0.452
& 0.522
& 8,987
& 479.98
& 4.01
& \underline{3.0}
& 14.03
& 0.418
& 0.606 \\

\midrule

\textbf{\oursmemupdater}
& $480\!\times\!832$
& \underline{0.142}
& \underline{0.823}
& \textbf{39}
& \textbf{14.11}
& \textbf{28.66}
& \textbf{735.3}
& \underline{17.52}
& \underline{0.526}
& \underline{0.537} \\

\midrule

\multicolumn{11}{c}{\textit{RealEstate10K (Out-of-domain)}} \\
\midrule

ReCoSplat
& $480\!\times\!480$
& \textbf{0.073}
& \textbf{0.934}
& \underline{1,237}
& \underline{117.93}
& \underline{11.69}
& \underline{3.1}
& \textbf{24.88}
& \textbf{0.869}
& \textbf{0.208} \\

ReCoSplat
& $480\!\times\!832$
& 0.576
& 0.568
& 5,818
& 554.85
& 2.22
& 2.9
& 13.56
& 0.622
& 0.640 \\

OF$^3$GS
& $405\!\times\!832$
& 0.487
& 0.634
& 6,649
& 355.07
& 4.80
& 3.0
& 14.15
& 0.590
& 0.566 \\

\midrule

\textbf{\oursmemupdater}
& $480\!\times\!832$
& \underline{0.125}
& \underline{0.866}
& \textbf{25}
& \textbf{8.94}
& \textbf{29.03}
& \textbf{1333.3}
& \underline{19.68}
& \underline{0.706}
& \underline{0.428} \\

\bottomrule
\end{tabular}
}
\end{table*}

\begin{figure}[t]
    \centering
   \vspace{-10pt}
   \includegraphics[width=\linewidth]{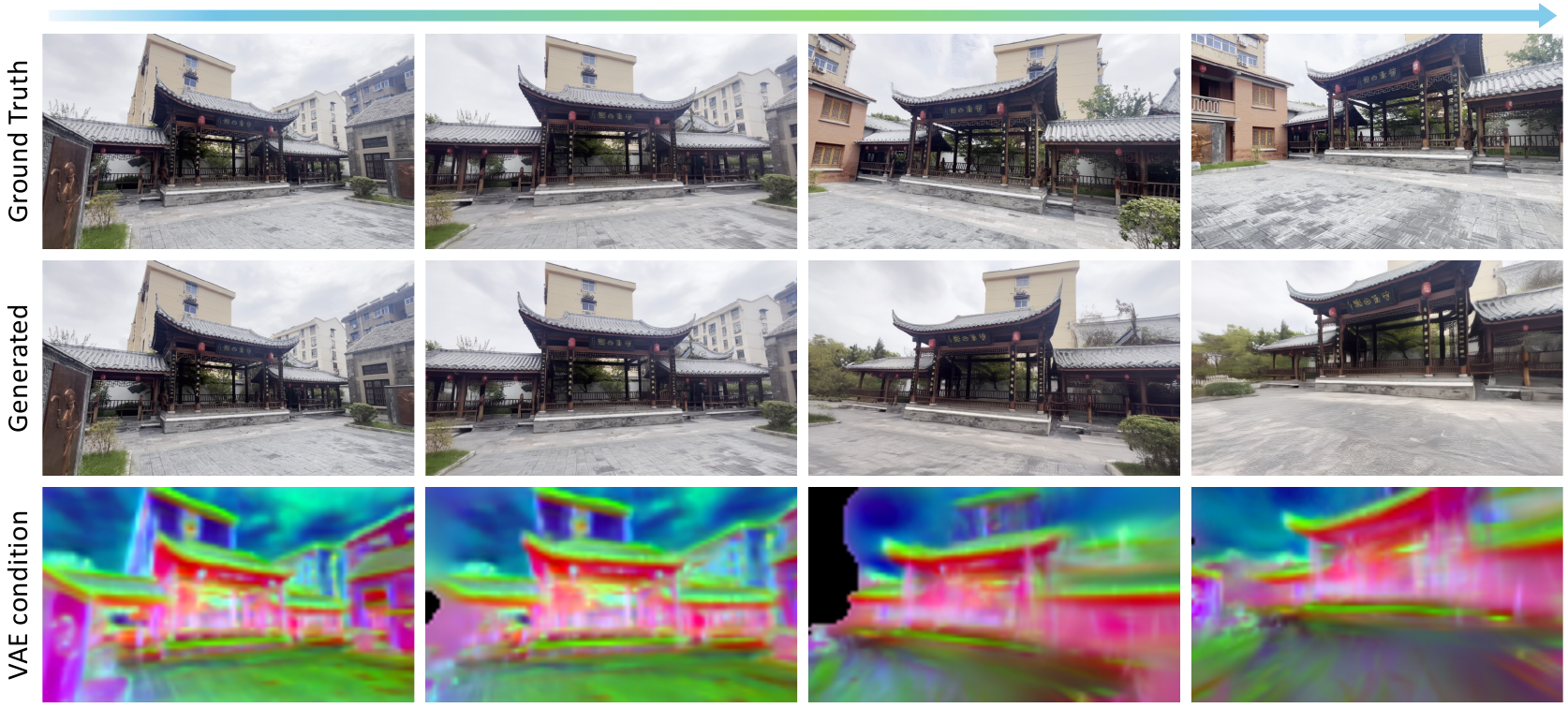}
    \vspace{-15pt}
    \caption{
    \textbf{Visualization of \oursmem conditioning for video generation.}
    We visualize the ground-truth and generated video, with target-view rendered VAE condition from the \oursmem, illustrating that spatially aligned latent memory provides consistent conditioning for video generation.
    }
    \label{fig:vae_condition_appendix}
\end{figure}

More importantly, the utility of this reconstruction is determined by how the memory conditions subsequent video generation. Fig.~\ref{fig:vae_condition_appendix} visualizes, for each target camera, the ground-truth view, the generated view, and the VAE condition directly rendered from \oursmem. The rendered latent remains spatially aligned with the target camera and preserves the structure of previously observed regions. By directly conditioning \oursvid on these spatially anchored VAE features, the model can generate novel content while remaining consistent with the persistent scene memory. Thus, \oursmem is not merely an efficient reconstruction module: it provides a compact, continuously updated, and directly renderable latent interface between the accumulated 3D scene and autoregressive video generation.

\section{Limitations and Future Work}

Despite its efficiency, \ours has several limitations that suggest directions for future work. First, \oursmem is primarily designed for static scenes and does not explicitly model independently moving objects. Extending the persistent memory to disentangle dynamic content from static geometry would enable more general interactive environments. Second, although the memory grows with explored spatial extent rather than generation length, its size can still increase under extremely large-scale exploration; adaptive pruning or hierarchical memory representations could further improve scalability. Finally, while the refinement mechanism can progressively correct inaccurate Gaussians, errors introduced during long-horizon autoregressive generation may still propagate to subsequent memory updates.

\end{document}